\documentclass{article} % For LaTeX2e
\usepackage{iclr2027_conference,times}

\iclrfinalcopy
\usepackage{amsmath,amsfonts,bm}

\def\eqref#1{equation~\ref{#1}}
\def\1{\bm{1}}

\DeclareMathAlphabet{\mathsfit}{\encodingdefault}{\sfdefault}{m}{sl}
\SetMathAlphabet{\mathsfit}{bold}{\encodingdefault}{\sfdefault}{bx}{n}

\usepackage{hyperref}
\usepackage{url}
\usepackage{booktabs}
\usepackage{graphicx}
\usepackage{wrapfig}
\usepackage[table]{xcolor}
\usepackage{caption}
\usepackage{graphicx} 
\usepackage{flafter}
\usepackage{placeins}
\usepackage{float}

\title{\ours: Video Diffusion Model is a Universal\\ Latent Action Learner}

\author{
\begin{tabular}{@{}l@{}}
Jisoo Kim$^{1*}$ \quad Chaehyun Kim$^{1*}$ \quad Junsu Kim$^{1}$ \quad Jiho Park$^{1}$ \quad Seongchan Kim$^{1}$ \\
\textbf{Eunbeen Hong$^{1}$ \quad Soowon Son$^{1}$ \quad Fu-En Yang$^{2}$ \quad Seungryong Kim$^{1}$}
\end{tabular} \\[4pt]
\normalfont \small $^{1}$KAIST \qquad $^{2}$NVIDIA \\[4pt]
\normalfont \textbf{Project page:} \href{https://cvlab-kaist.github.io/DreamX}{\texttt{https://cvlab-kaist.github.io/DreamX}}
}

\newcommand{\paragrapht}[1]{\vspace{-10pt}\paragraph{#1}}
\newcommand{\ours}{DreamX}

\begin{document}

\maketitle
\fancyhead{}

% Keep the teaser below the title and above the abstract, without floating.
\begin{figure}[H]
    \centering
    \vspace{-30pt}
    \includegraphics[width=\linewidth]{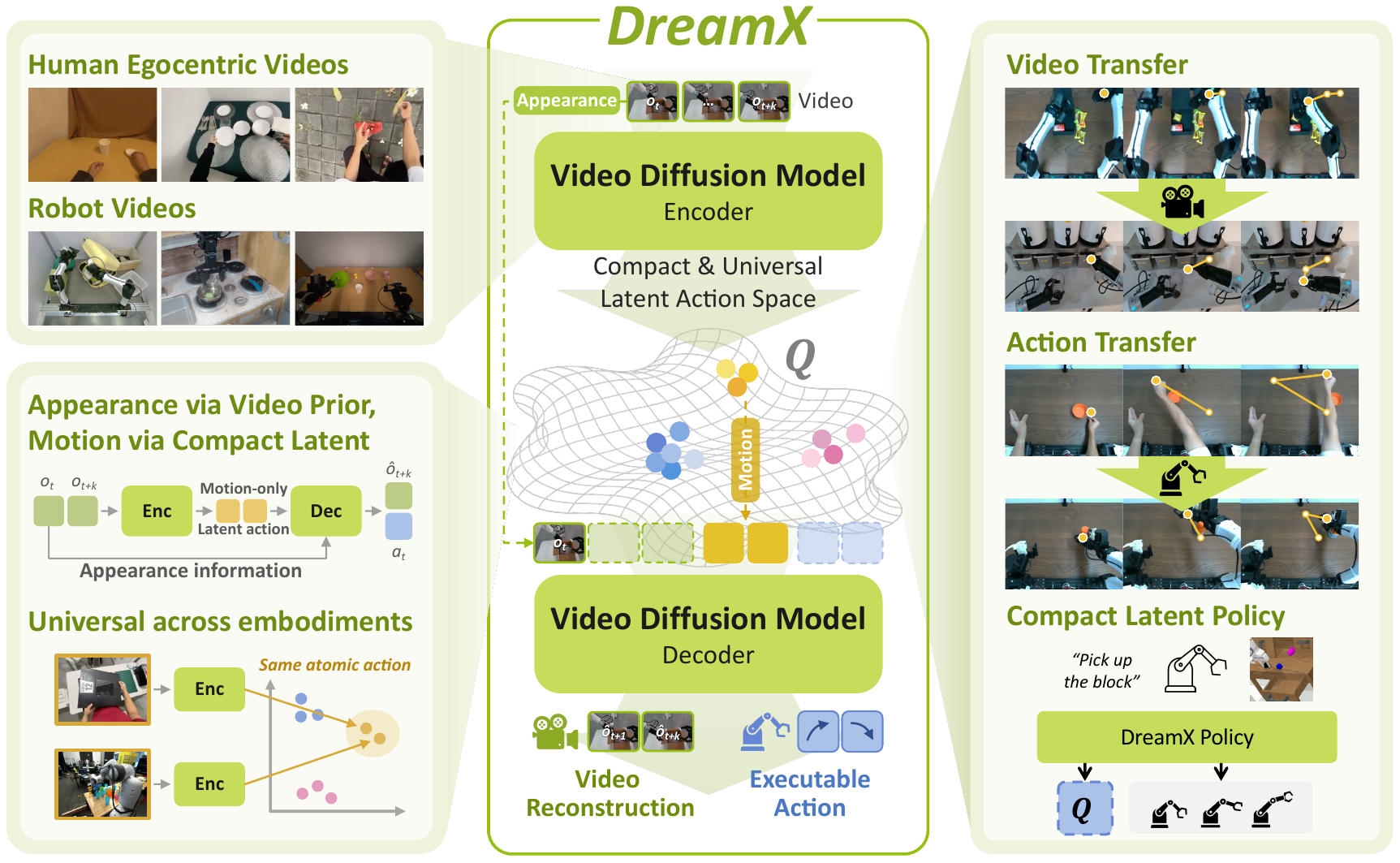}
    \vspace{-18pt}
    \caption{\textbf{DreamX learns a shared action interface from egocentric human videos and robot videos.}
    Compact, embodiment-agnostic latent actions extracted from a pretrained video diffusion model capture motion consistently across changes in embodiment and scene appearance. The same representation enables transferring motion from a source video to a target video, decoding human motion into executable robot actions without paired human--robot training data, and training a lightweight latent policy that jointly predicts latent actions and robot actions.}
\label{fig:teaser}
    \end{figure}\vspace{-5pt}

\begin{abstract}
% Learning a universal action representation from video requires capturing the dynamics shared across different realizations of an action while remaining invariant to changes in embodiment and appearance. Existing latent action models (LAMs) derive actions from visual transitions, but their representations may entangle the underlying action with incidental visual variations, limiting transfer across embodiments. In this paper, we introduce \emph{\ours}, a framework that directly leverages the rich representation of pre-trained video diffusion model and turns into a universal latent action learner. Our model directly reads out compact latent actions from spatiotemporal representations, compressing each latent frame by 300$\times$, and leverages the generative prior to separate action-relevant dynamics from visual appearance. The resulting latent space provides a shared action interface across human and robot embodiments, environments, and action parameterizations. Extensive experiments demonstrate state-of-the-art representation quality on LARY-Bench and enable any-to-any video transfer across embodiments and environments. Moreover, our approach achieves human-to-robot policy transfer using only human video demonstrations for task-specific adaptation. We further show that this latent representation can be a basis for compact policy model, which dramatically reduce the video diffusion model's computational burden and enhance the performance at the same time.

Learning a universal action representation from video requires modeling dynamics that are consistent across different executions of the same action yet invariant to changes in embodiment and appearance. Existing latent action models infer actions from visual transitions, but their representations often entangle the underlying action with irrelevant visual variations, limiting transfer across embodiments. We introduce \textbf{\ours}, which repurposes a pretrained video diffusion model as a universal latent action learner. Our approach extracts compact latent actions from the video model's rich spatiotemporal features, compressing each latent video frame by $\boldsymbol{300\times}$ ($\boldsymbol{1{,}200 \rightarrow 4}$ tokens). In addition, by leveraging the video model's generative prior, these latent actions capture action-relevant dynamics disentangled from visual appearance. The resulting latent space serves as a shared action interface across embodiments (human and robot), environments, and action parameterizations. On LARY-Bench, \ours{} achieves state-of-the-art performance among latent action models. Moreover, actions encoded in one setting remain effective in another: the latent space supports any-to-any video transfer and zero-shot action transfer across diverse embodiments and environments. Furthermore, this shared interface also enables policy learning in a compact latent action space, reducing the computational cost of video-diffusion-based policies.

% Learning a universal action representation from video requires capturing dynamics shared across executions of an action despite changes in embodiment and appearance. Existing latent action models (LAMs) infer actions from visual transitions, but often conflate action with incidental visual variation, limiting transfer across embodiments. We introduce \emph{\ours}, which transforms a pretrained video diffusion model into a universal latent action learner. Our method extracts compact latent actions from its rich spatiotemporal features, compressing each latent video frame by 300$\times$, while using the model's generative prior to disentangle action-relevant dynamics from visual appearance. The resulting latent space provides a shared action interface across human and robot embodiments, environments, and action parameterizations. On LARY-Bench, \emph{\ours} achieves state-of-the-art representation quality. More importantly, the same actions support any-to-any video transfer across embodiments and environments, as well as zero-shot cross-embodiment action transfer. A compact latent policy built on this space further improves task performance while reducing the computational cost of video-diffusion-based control.
\end{abstract}

\section[Introduction]{Introduction}
\label{sec:intro}

Generalist robot policies, including vision-language-action (VLA) models~\citep{kim24openvla,bjorck2025gr00t,intelligence2025pi0} and world action models (WAMs)~\citep{kim2026cosmos,ye2026world}, benefit from diverse robot demonstrations, but collecting such data at scale remains costly.
Human egocentric videos offer a promising alternative as a scalable source of diverse manipulation behaviors, and recent works leverage these videos for policy learning~\citep{li2025h2r,kareer2025egomimic}.
However, action supervision remains tied to the embodiment that produces it.
Human videos lack robot action labels, and even robot demonstrations use different control spaces across embodiments.
The same behavior can therefore correspond to different low-level actions, making action labels difficult to share directly across embodiments.
Overcoming this mismatch calls for a common action representation that preserves shared motion and effects across embodiments.

% P3->2: LAMs → visual transition as common action interface, but is it really action-centric?

Latent action models (LAMs) offer a promising route toward learning such a representation from video.
Although humans and robots have different bodies and control spaces, their actions can all be observed as changes in a visual scene.
LAMs typically use dedicated encoders to compress these changes into compact representations, called latent actions~\citep{ye2025latentactionpretrainingvideos,chen2025moto,bu2025univla}, where these encoders are trained using a decoder that reconstructs future frames solely from the current frame and latent actions. However, reconstruction accuracy alone does not ensure that latent actions isolate the underlying action. The learned representations can also encode domain-specific visual information, such as viewpoint and scene appearance, whenever it helps reconstruction~\citep{nikulin2025latent}. Capturing action-relevant structure shared across embodiments while separating it from domain-specific visual factors therefore remains a challenge.
% These encoders are trained through future-frame reconstruction: a decoder predicts subsequent frames from the latent actions and current frame.
% However, future-frame reconstruction does not explicitly constrain latent actions to represent action alone. To reconstruct the full visual transition, existing LAMs also encode domain-specific visual information associated with viewpoint, scene appearance, and embodiment.
% These encoders are often trained using a decoder that reconstructs future frames solely from the current frame and latent actions. However, this objective only requires accurate reconstruction; it does not require latent actions to isolate the underlying action. As a result, latent actions in existing LAMs can also encode domain-specific visual information, such as viewpoint, scene appearance, and embodiment, whenever it helps reconstruction.~\citep{nikulin2025latent}
% However, visual change is not uniquely determined by action.
% Viewpoint, scene appearance, and embodiment also shape the observed transition, so latent actions can capture domain-specific visual factors alongside the underlying action.
% Capturing action-relevant structure shared across embodiments while separating it from domain-specific visual factors therefore remains a challenge.

% P4->3: VDM → rich dynamics already exist, but how do we extract action from them?

Pretrained video diffusion models provide a promising foundation for addressing this challenge.
First, through large-scale video generation, they learn rich spatiotemporal representations that capture motion, correspondence, geometry, and semantic structure~\citep{difftrack,wiedemer2025videomodelszeroshotlearners,son2025probing}.
Such representations encode how scenes, objects, and agents evolve over time, making them a natural source of action-relevant dynamics.
Second, these models possess a generative prior that enables future-frame synthesis conditioned on an initial frame and an additional signal.
With scene appearance supplied by the initial frame, additional conditioning can guide how the scene evolves. Together, these priors motivate extracting latent actions directly from pretrained spatiotemporal representations and supervising their learning with a pretrained generative decoder.
% Together, these priors raise the question of whether latent actions can be read out directly from the video model's spatiotemporal representations while disentangling action-relevant dynamics from visual appearance.

% P5->4: Ours

% We introduce \textbf{\ours}, a framework for learning universal latent actions across embodiments by leveraging the priors of a pretrained video diffusion model.
% Our key idea is to use its spatiotemporal representations to encode action-relevant dynamics and its generative prior to supervise latent-action learning.
% To this end, we introduce learnable query tokens alongside video tokens and jointly encode them through the model's spatiotemporal attention.
% The queries read out dynamics across frames from the pretrained representations, condensing them into compact latent actions with $\mathbf{300\times}$ fewer tokens than the video representation.
% We then use these latent actions, rather than text, to condition the same model to generate future frames from the initial frame.
% Given scene appearance from the initial frame and the pretrained generative prior, the latent actions are encouraged to encode the motion that leads to the observed future.
We introduce \textbf{\ours}, a framework for learning universal latent actions across embodiments using a pretrained video diffusion model. We pass learnable query tokens alongside the video tokens through the model's spatiotemporal attention, allowing the queries to gather action-relevant information across frames.
The queries condense frame-to-frame changes represented by \textbf{1,200} frame tokens into just \textbf{4} latent-action tokens, yielding a $\mathbf{300\times}$ reduction in token count. These latent actions replace text conditioning in the same model to reconstruct future frames from the initial frame. With scene appearance supplied by the initial frame and visual detail by the pretrained generative prior, the reconstruction objective encourages latent actions to capture the motion needed to explain the observed future.

% With scene appearance supplied by the initial frame and the pretrained generative prior, the latent actions are encouraged to capture the motion needed to explain the observed future.
% Building on these latent actions, we develop a latent-action policy that predicts compact latent-action codes instead of future videos, reducing the computational burden of video-based policy models.
% Our experiments show that \ours{} learns universal latent actions, enabling \emph{any-to-any video transfer} and \emph{human-to-robot action transfer} without paired training data.
% The latent-action policy further benefits from human video demonstrations, and \ours{} achieves state-of-the-art performance among latent action models on LARYBench~\citep{nie2026lary}.
Trained on heterogeneous human and robot videos, \ours{} achieves state-of-the-art performance among latent action models on LARYBench~\citep{nie2026lary} and enables \emph{any-to-any video transfer} and \emph{human-to-robot action transfer} without paired training data. As a downstream application, we also show that the learned latent actions can serve as prediction targets for an observation-conditioned robot policy, without requiring future-video generation during execution. Together, these results suggest that \ours{} learns a shared, transferable latent action space that supports motion transfer across embodiments and can also serve as a basis for downstream policy learning.

In summary, our contributions are as follows: 
% (1) We introduce \ours{}, a framework that leverages the spatiotemporal and generative priors of a pretrained video diffusion model to learn universal latent actions. 
% (2) By leveraging the pretrained spatiotemporal prior, \ours{} encodes motion-relevant structure into latent actions, achieving state-of-the-art performance among latent action models on LARYBench, with analyses revealing the motion information captured by the queries. 
% (3) By exploiting priors of pretrained video diffusion models, \ours{} learns universal latent actions that support \emph{any-to-any video transfer} and \emph{human-to-robot action transfer} without paired source--target data, as well as policy learning from human and robot videos.
\begin{itemize}
    \item We introduce \ours{}, a framework that uses the spatiotemporal and generative priors of a pretrained video diffusion model to learn universal latent actions. The resulting representations achieve state-of-the-art performance among latent action models on LARYBench.
    \item We demonstrate that the learned latent space supports \emph{any-to-any video transfer} and \emph{human-to-robot action transfer} without paired source--target training data. We further show the use of compact latent-action codes as prediction targets for downstream policy learning.
\end{itemize}
\section{Related Works}
\label{sec:relwork}

\paragraph{Latent action models.}
Latent action models (LAMs) learn action representations from visual transitions, enabling action-unlabeled videos to supervise downstream control. LAPO~\citep{schmidt2024learning} and Genie~\citep{bruce2024genie} established the paradigm, recovering discrete latent actions through an inverse--forward dynamics bottleneck over adjacent frames. LAPA~\citep{ye2025latentactionpretrainingvideos} further scales this recipe for vision-language-action pretraining, and Moto~\citep{chen2025moto} tokenizes visual motion for manipulation. Later work incorporates additional cues: UniVLA~\citep{bu2025univla} grounds latent actions with semantic visual features, villa-X~\citep{chen2025villaxenhancinglatentaction} integrates language and robot-state dynamics, and CoMo~\citep{yang2026como} and DiLA~\citep{zhang2026dila} separate motion from visual content. Across these variants, the action space is learned from scratch on a local visual transition, often a single frame pair. Such supervision offers no view of how motion persists beyond one step, so the encoder entangles action with appearance and yields representations tied to their training embodiments.

\paragrapht{Emergent capabilities of video generative models.}

Video generative models acquire spatiotemporal structure by modeling how scenes evolve over time. Their internal representations support correspondence, point tracking, optical flow, and geometric understanding without task-specific supervision~\citep{difftrack, vélez2025imagevideoempiricalstudy, vedder2024zeroflowscalablesceneflow, huang20253dvideofoundationmodels}, and text-conditioned models further exhibit semantic grounding over objects, behaviors, and interactions~\citep{wiedemer2025videomodelszeroshotlearners}. Rather than encoding physical laws explicitly, these representations capture motion, geometry, and semantics jointly: which entities are present, where they are, and how they move over time. This is precisely the composition a latent action must express, which is a structured change of a specific entity in a specific place, and it is acquired from web-scale video rather than from a single frame pair.

\paragrapht{Video generative priors for latent action learning.}

Recent work connects these two lines. VPP~\citep{hu2024video} conditions a manipulation policy on the predictive visual representations inside a fine-tuned video diffusion model, but still learns the policy from action labels. AdaWorld~\citep{gao2025adaworld} learns latent actions for adaptable video prediction, and VideoWorld2~\citep{videoworld2} pairs a pretrained video diffusion model for generation with a separate latent dynamics model. In both, the video model acts as a \emph{generative} prior supplying pixels or rollouts, while the action representation is produced by an external module trained from scratch on frame-level transitions. Thus, the latent action encoder does not directly read the video diffusion backbone’s internal representations. We instead read latent actions directly from the model's rich spatiotemporal representation, making maximal use of video pretraining and requiring no per-embodiment action encoder.

\section[Method]{Method}
\label{sec:method}

\begin{figure}[t]
    \centering
    \includegraphics[width=\linewidth]{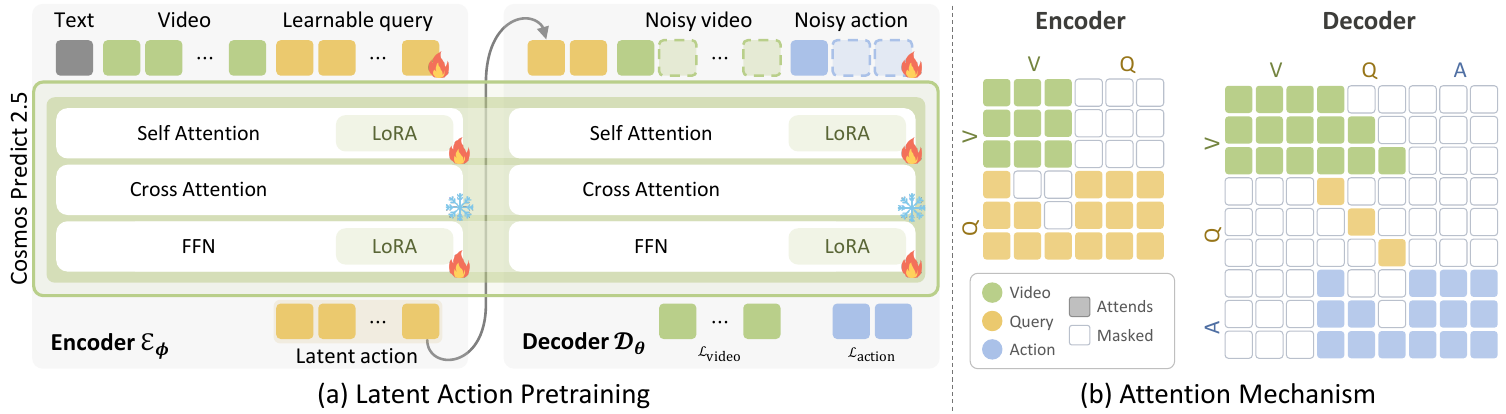}
    \vspace{-15pt}
    % Previous caption, preserved before revision:
    % \caption{Overview of \textbf{\ours}. A pretrained video foundation model is adapted to encode local motion into latent-action queries. The frozen representation is distilled into compact downstream interfaces, which support video transfer and low-level action generation across embodiments.\textcolor{red}{Place holder}}
    \caption{\textbf{\ours{} architecture and attention masks.} (a) Learnable queries read out the pretrained video backbone's spatiotemporal representations to encode latent actions. A decoder initialized from the same pretrained backbone reconstructs future video from the latent actions and the initial frame, with auxiliary action supervision when labels are available. (b) Encoder and decoder attention masks control information flow among video, query, and action tokens. In the decoder, video and action tokens read the latent-action queries but do not attend to one another.}
    \label{fig:pipeline}
    \vspace{-12pt}
\end{figure}

\label{sec:method_overview}

Given a video and instruction, \ours{} uses learnable queries to extract compact latent actions from a pretrained video diffusion backbone's spatiotemporal features. Future-frame reconstruction conditioned on these latent actions and the initial frame provides generative supervision (Sec.~\ref{sec:pretraining}). The learned representation enables video and action transfer with a target initial frame or robot state (Sec.~\ref{sec:transfer}). For control, the \ours{} policy predicts compact codes and low-level actions without generating future video (Sec.~\ref{sec:latentwam}).
 
\subsection{Preliminaries: Video Diffusion Models}
\label{sec:prelim}
 
We build on Cosmos-Predict2.5~\citep{cosmos}, a pretrained video diffusion transformer. A causal video autoencoder compresses a clip into $N$ latent frames, $\mathbf{V}=(V_0,\ldots,V_{N-1})$, where $V_0$ encodes only the initial frame. These frames are patchified into spatial tokens and jointly modeled through spatiotemporal attention. Through video pretraining, the model learns spatiotemporal priors that capture spatial context and temporal relationships across the clip.
Conditioned on an initial frame and an additional signal such as text, collectively denoted by $c$, the model learns future-video generation through flow matching. Given clean video latents $x$ and noise $\epsilon \sim \mathcal{N}(\mathbf{0},\mathbf{I})$, we form $x_\tau=(1-\tau)x+\tau\epsilon$, with $\tau \in [0,1]$, and minimize
\begin{equation}
\mathcal{L}_{\mathrm{video}}
=
\mathbb{E}_{x,c,\epsilon,\tau}
\left[
\left\lVert
v_\theta(x_\tau,\tau;c)-(\epsilon-x)
\right\rVert_2^2
\right],
\label{eq:fm-loss}
\end{equation}
where $v_\theta$ predicts the flow velocity.

By optimizing this objective over large-scale video data, the model acquires a generative prior over scene appearance and plausible dynamics. When the initial frame supplies the scene's appearance, additional conditioning steers how the scene evolves. We exploit this property by using latent actions to steer future-frame reconstruction, whose objective supervises what the latent actions encode.

% \subsection{Latent Action Pretraining}
\subsection{Learning Universal Latent Actions from Video Diffusion Priors}
\label{sec:pretraining}
 
\ours{} adapts a pretrained video diffusion backbone for two complementary roles: encoding observed dynamics into latent actions and supervising them through future-frame reconstruction to encourage separation of motion from visual appearance. Building on Sec.~\ref{sec:prelim}, we introduce learnable queries that read out the video representations to summarize observed dynamics as latent actions, while the decoder reconstructs the future from these latent actions and the initial frame. Leveraging a video diffusion backbone for both the encoder and decoder allows its generative prior to account for visual appearance, encouraging the latent action tokens to focus on motion.

\begin{figure}[t]
    \centering
    \includegraphics[width=\linewidth]{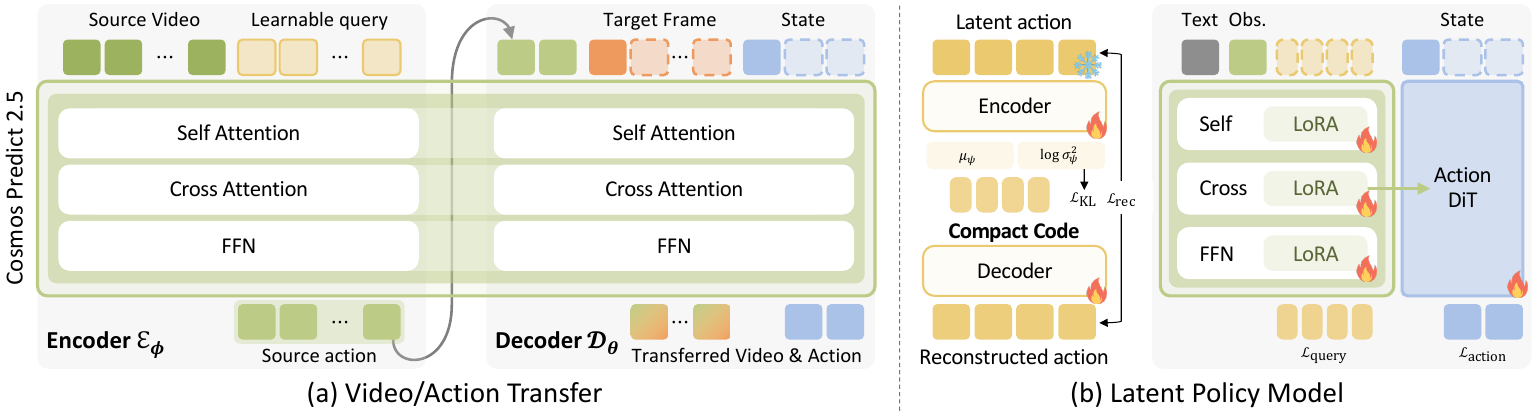}
    \vspace{-18pt}
    % Previous caption, preserved before revision:
    % \caption{Downstream adaptation of \textbf{\ours}. The pretrained latent-action model is kept frozen. (a) \textbf{Transfer.} Latent actions extracted from a source video condition the pretrained decoder, which renders the source motion from the first frame of a different scene and decodes low-level actions for a target robot. (b) \textbf{Latent action policy.} A lightweight bottleneck compresses the latent actions into a compact continuous code, which the video model learns to generate from the current observation and instruction, while a separate action head decodes low-level actions from the code.}
    \caption{\textbf{Downstream applications of \ours{}.} (a) \textbf{Motion transfer.} Source latent actions condition the \ours{} decoder to generate video from a target initial frame or low-level actions from a target robot state, without paired source--target training data. (b) \textbf{\ours{} policy.} A variational bottleneck compresses latent actions into compact codes while preserving the token layout. Conditioned on the current observation and instruction, the policy generates these codes alongside low-level actions through a separate DiT action head.}
    \vspace{-10pt}
    \label{fig:downstream}
\end{figure}

\paragrapht{Latent Action Encoding from Spatiotemporal Dynamics.}

\ours{} uses pretrained video model itself to encode latent actions. As shown in Figure~\ref{fig:pipeline}(a), we append learnable query tokens $\mathbf{q}$ to the video token sequence and process them jointly through the backbone's spatiotemporal attention. We assign a group of $K$ query tokens $\mathbf{q}_g$ to each latent frame $V_g$. Once contextualized by the video, their output features form the latent-action group $Q_g$. All $N$ groups are extracted in a single forward,
\begin{equation}
    Q = \mathcal{E}_\phi(\mathbf{q};\mathbf{V},\ell),
    \label{eq:encode}
\end{equation}
where $\mathcal E_\phi(\cdot)$ is the encoder and $\ell$ is the language instruction supplied through the backbone's original cross-attention. As shown in Figure~\ref{fig:pipeline}(b), query-to-video attention is frame-causal, so $Q_g$ directly attends to $V_{0:g}$, while query-to-query attention is unrestricted. This lets the queries encode temporally grounded motion from the pretrained dynamics, with query-layout settings in Appendix~\ref{app:pretraining_details}.

\paragrapht{Generative Decoding for Action-Centric Latent Learning.}

Encoding from video alone may retain scene- and embodiment-specific visual cues. We therefore use the same pretrained video backbone as the encoder to supervise latent actions through future-frame reconstruction, conditioned on $V_0$ and $Q_{0:N-1}$ without text. As $V_0$ provides scene appearance and the pretrained generative prior captures plausible evolution, $Q$ is encouraged to encode the motion needed to explain the future, rather than visual cues already available to the decoder.

As shown in Figure~\ref{fig:pipeline}(a), the decoder $\mathcal D_\theta(\cdot)$ generates future frames conditioned on $Q_{0:N-1}$, concatenated with the video tokens along the temporal sequence. When action labels are available, it also takes the proprioceptive state $p_0$ and decodes action chunks as an auxiliary task:
\begin{equation}
    \hat{V},\hat{A} = \mathcal{D}_\theta(Q_{0:N-1},V_0,p_0).
\end{equation}
Under the decoder attention mask in Figure~\ref{fig:pipeline}(b), latent-action queries serve solely as conditioning signals and do not attend to any tokens. The video and action streams attend to these queries without attending to each other. Specifically, video frame $V_t$ attends to the query prefix $Q_{0:t}$, while action chunk $A_m$ attends to $Q_{0:m}$. 
\paragrapht{Training Objective.}
We train both the encoder and the decoder end-to-end with the standard flow-matching objective applied to video and action generation as Eq.~\ref{eq:stage-a-loss}, with $\lambda = 0$ for video clips without action labels:
\begin{equation}
    \mathcal{L} = \mathcal{L}_{\text{video}} + \lambda\,\mathcal{L}_{\text{action}}.
    \label{eq:stage-a-loss}
\end{equation}

\subsection{Motion Transfer via Latent Actions}
\label{sec:transfer}

Building upon the rich generative prior from the video diffusion backbone, \ours{} learns latent actions that capture motion transferable across embodiments, even though cross-embodiment transfer is not an explicit training objective. This capability emerges from the pretraining described in the previous section and requires no additional training. The auxiliary robot action branch further encourages the latent actions to encode action-relevant cues and further enables motion extracted from a source video to be decoded into continuous actions executable by a target embodiment.

Given a source video, we extract its latent actions using the frozen \ours{} encoder,
$Q^{\mathrm{src}}=\mathcal{E}_\phi(\mathbf{q};\mathbf{V}^{\mathrm{src}},\ell^{\mathrm{src}})$,
which captures the motion to be transferred. As shown in Figure~\ref{fig:downstream}(a), we reuse the same $Q^{\mathrm{src}}$ in two complementary ways. For \emph{video transfer}, we combine it with a target initial frame $V^{\mathrm{tgt}}_0$ and use the frozen \ours{} decoder $\mathcal{D}_\theta$ to generate a target video that reproduces the source motion under the target scene and embodiment. For \emph{action transfer}, we instead combine $Q^{\mathrm{src}}$ with the target robot state $p^{\mathrm{tgt}}_0$ and decode the same motion into executable low-level actions,
\begin{equation}
    \hat{V}^{\mathrm{tgt}} = \mathcal{D}_\theta\big(Q^{\mathrm{src}}, V^{\mathrm{tgt}}_0\big),
    \qquad
    \hat{A}^{\mathrm{tgt}} = \mathcal{D}_\theta\big(Q^{\mathrm{src}}, p^{\mathrm{tgt}}_0\big).
    \label{eq:motion-transfer}
\end{equation}
Thus, the shared latent action realizes the same underlying motion either as a generated video in a new domain or as executable actions on a target robot, \emph{without paired source--target datasets}.

% Required packages:
% \usepackage{booktabs}
% \usepackage{graphicx}
% \usepackage[table]{xcolor}

\begin{table*}[t]
\centering
\caption{\textbf{LARYBench action regression and classification.}
Regression reports normalized action MSE and classification reports accuracy.
\textbf{Bold} and \underline{underline} denote the best and second-best results.}
\vspace{-0.8em}
\label{tab:lary_regression_classification}

\resizebox{\textwidth}{!}{%
\begin{tabular}{lccccc|cccc}
\toprule
& \multicolumn{5}{c}{\textbf{Action regression $\downarrow$}}
& \multicolumn{4}{c}{\textbf{Action classification $\uparrow$}} \\
\cmidrule(lr){2-6}
\cmidrule(lr){7-10}
Method
& CALVIN
& VLABench
& RoboCOIN
& AgiBotWorld-Beta
& Avg.
& Atomic robot
& Composite human
& Composite robot
& Avg. \\
\midrule

\multicolumn{10}{l}{\textit{Latent actions from 2 frames}} \\

LAPA
& 0.96 & 0.95 & 0.96 & 1.00 & 0.97
& 22.27 & 14.61 & 23.64 & 20.17 \\

UniVLA
& 0.82 & 0.74 & 0.94 & 0.97 & 0.87
& 18.62 & 19.08 & 18.56 & 18.75 \\

villa-X
& 0.86 & 0.72 & 0.94 & 0.97 & 0.87
& 15.00 & 17.80 & 29.90 & 20.90 \\

LAPA-MAGVIT2
& 0.59 & 0.36 & 0.80 & 0.83 & 0.65
& 33.12 & 59.70 & 29.53 & 40.78 \\

LAPA-SigLIP2
& 0.57 & 0.30 & 0.86 & 0.88 & 0.65
& 46.83 & 54.74 & 29.44 & 43.67 \\

LAPA-DINOv2
& 0.55 & 0.26 & 0.85 & 0.86 & 0.63
& 58.04 & 55.86 & 34.19 & 49.36 \\

LAPA-DINOv3
& 0.50 & 0.25 & 0.82 & 0.84 & 0.60
& 56.27 & 64.19 & 27.04 & 49.17 \\

% MoTo-endpoint~\citep{chen2025moto}
% & 0.45 & 0.19 & 0.62 & 0.75 & 0.50
% & 41.57$^{\dagger}$ & 38.56 & 41.88 & 40.67 \\

\midrule
\multicolumn{10}{l}{\textit{Latent actions from video}} \\

MoTo~\citep{chen2025moto}
& 0.43 & 0.13 & 0.54 & 0.68 & 0.44
& 41.57 & 55.11 & 56.34 & 51.01 \\

Adaworld~\citep{gao2025adaworld}
& 0.64 & 0.26 & 0.94 & 0.94 & 0.69
& 31.58 & 49.8 & 51.48 & 44.29 \\

VideoWorld2~\citep{ren2026videoworld2learningtransferable}
& 0.72 & 0.41 & 0.82 & 0.85 & 0.70
& 14.50 & 30.24 & 37.36 & 27.37 \\

% CoMo~\citep{yang2026como}
% & -- & -- & -- & -- & --
% & -- & -- & -- & -- \\

DiLA~\citep{zhang2026dila}
& 0.43 & 0.09 & 0.51 & 0.52 & 0.38
& 58.51 & 65.47 & 62.76 & 62.25 \\

\textbf{DreamX (Ours)}
& \textbf{0.15} & \textbf{0.05} & \textbf{0.26} & \textbf{0.22} & \textbf{0.17}
& \textbf{68.03} & \textbf{72.72}
& \textbf{68.26} & \textbf{69.67} \\

\textbf{DreamX (Ours) + FAE-32}
& \underline{0.23} & \underline{0.07} & \underline{0.31} & \underline{0.26} & \underline{0.22}
& \underline{65.90} & \underline{69.49} & \underline{66.06} & \underline{67.15} \\

\bottomrule
\end{tabular}%
}

\vspace{3pt}
\begin{minipage}{0.98\textwidth}
\end{minipage}
\vspace{-15pt} % Adjust the gap between Table 1 and the following text.
\end{table*}

\subsection{Policy Learning with Compact Latent Actions}
\label{sec:latentwam}
% For control, we repurpose the same pretrained video backbone used by \ours{} to predict latent actions in place of future video. Although $Q$ is compact in token count, each token retains the full backbone width, making direct diffusion high-dimensional~\citep{dcae15}. We therefore compress $Q$ into a low-dimensional code $Z$ and train the policy model to predict $Z$ from the current observation and instruction. As a result, the latent space can also work as a basis of latent world action model model to predict the action given the observation.~\citep{kim2026compact}
As a downstream application, we examine whether the latent actions learned by \ours{} can serve as prediction targets for an observation-conditioned robot policy. Whereas motion transfer uses latent actions extracted from a source demonstration, the policy predicts them from the current observation and instruction. To this end, we repurpose the same pretrained video backbone used by \ours{} to predict latent actions in place of future video~\citep{kim2026compact}. Although $Q$ is compact in token count, each token retains the full backbone width, making direct diffusion high-dimensional~\citep{dcae15}. We therefore compress $Q$ into a low-dimensional code $Z$ first, which serves as the policy's prediction target.

\paragrapht{Latent Action Compression.}
% Following~\citep{fae}, we learn a variational bottleneck to compress each token in $Q$ from backbone width $D$ to $d \ll D$ preserving the token layout, yielding a compact denoising target for latent policy learning, as shown in Figure~\ref{fig:downstream}(b):
% \begin{equation}
%     Z = \mu_\psi(Q) + \sigma_\psi(Q) \odot \epsilon,
%     \qquad
%     \epsilon \sim \mathcal{N}(\mathbf{0}, \mathbf{I}),
%     \label{eq:fae-encode}
% \end{equation}
% where $\big(\mu_\psi(Q),\,\log \sigma_\psi^2(Q)\big)=f_\psi(Q)$ parameterizes the factorized Gaussian posterior $q_\psi(Z \mid Q)$ and $\hat{Q}=g_\psi(Z)$ reconstructs the original latent actions. Only $f_\psi$ and $g_\psi$ are trained with
% \begin{equation}
%     \mathcal{L}_{\mathrm{FAE}}
%     = \big\lVert \hat{Q} - Q \big\rVert_2^2
%     + \beta\, \mathrm{KL}\big(q_\psi(Z \mid Q) \,\Vert\, \mathcal{N}(\mathbf{0}, \mathbf{I})\big).
%     \label{eq:fae-objective}
% \end{equation}
% The reconstruction term encourages the codes to retain the information in $Q$, while the KL term regularizes their posterior toward a standard Gaussian.
Following~\citep{fae}, we learn a variational bottleneck that compresses each latent-action token from backbone width $D$ to $d \ll D$ while preserving the token layout, as shown in Figure~\ref{fig:downstream}(b). With the \ours{} encoder frozen, we train the bottleneck using feature reconstruction and KL regularization. After training, we use its posterior mean to obtain deterministic compact codes $Z$ for policy learning. The formulation and implementation details are provided in Appendix~\ref{app:fae_details}.

\paragrapht{Policy Learning on Compact Latent Action Space.}
With the learned bottleneck frozen, the compact codes serve as prediction targets for the \ours{} policy $\pi(\cdot)$. As shown in Figure~\ref{fig:downstream}(b), the policy generates these codes from the current frame $V_0$ and instruction $\ell$, while a separate DiT action head predicts low-level actions conditioned on the generator's latent-action features, the same observation and instruction, and the state $p_0$:
\begin{equation}
    \pi(Z,A \mid V_0,\ell,p_0).
    \label{eq:latentwam-policy}
\end{equation}
Action tokens attend to latent-action features, but not vice versa. This separation allows human videos to supervise latent-code prediction without robot action labels, while robot demonstrations additionally supervise low-level action prediction. We train the policy with
\begin{equation}
    \mathcal{L}_{\mathrm{policy}} = \mathcal{L}_{\mathrm{latent}} + \lambda\,\mathcal{L}_{\mathrm{action}},
    \label{eq:policy-loss}
\end{equation}
where both terms follow Eq.~\ref{eq:fm-loss}, applied to compact codes $Z$ and low-level actions $A$ respectively, with $\lambda=0$ for human videos without action labels. Further details are provided in Appendix~\ref{app:policy_implementation}.

\section{Experimental Results}
\label{sec:experiment}

We evaluate whether \ours{} learns universal latent actions by examining how well they capture motion (Sec.~\ref{sec:representation_quality}) and whether they can transfer across scenes and embodiments (Secs.~\ref{sec:video_transfer}--\ref{sec:action_transfer}). We further assess their utility for learning effective and efficient policies from human and robot demonstrations (Sec.~\ref{sec:policy_eval}), followed by ablations and analysis of the learned representation (Secs.~\ref{sec:ablation}--\ref{sec:analysis}).

\paragrapht{Implementation details.}
We instantiate \ours{} with Cosmos-Predict2.5 2B and pretrain on robot demonstrations from MolmoAct2-BimanualYAM~\citep{fang2026molmoact2}, Bridge~\citep{walke2023bridgedata}, DROID~\citep{khazatsky2024droid}, and AgiBot~\citep{bu2025agibot}, together with human videos from EgoDex~\citep{hoque2026egodex} and EgoVerse~\citep{punamiya2026egoverse}. Videos are processed at $480\times640$ resolution with variable-length clips of up to 29 frames. We assign four query tokens to each VAE latent frame. We adapt the encoder and decoder using separate rank-16 LoRA modules for 78,000 training steps. Further training details are provided in Appendix~\ref{app:pretraining_details}.

\begin{figure*}[t]
    \centering
    \includegraphics[width=\textwidth]{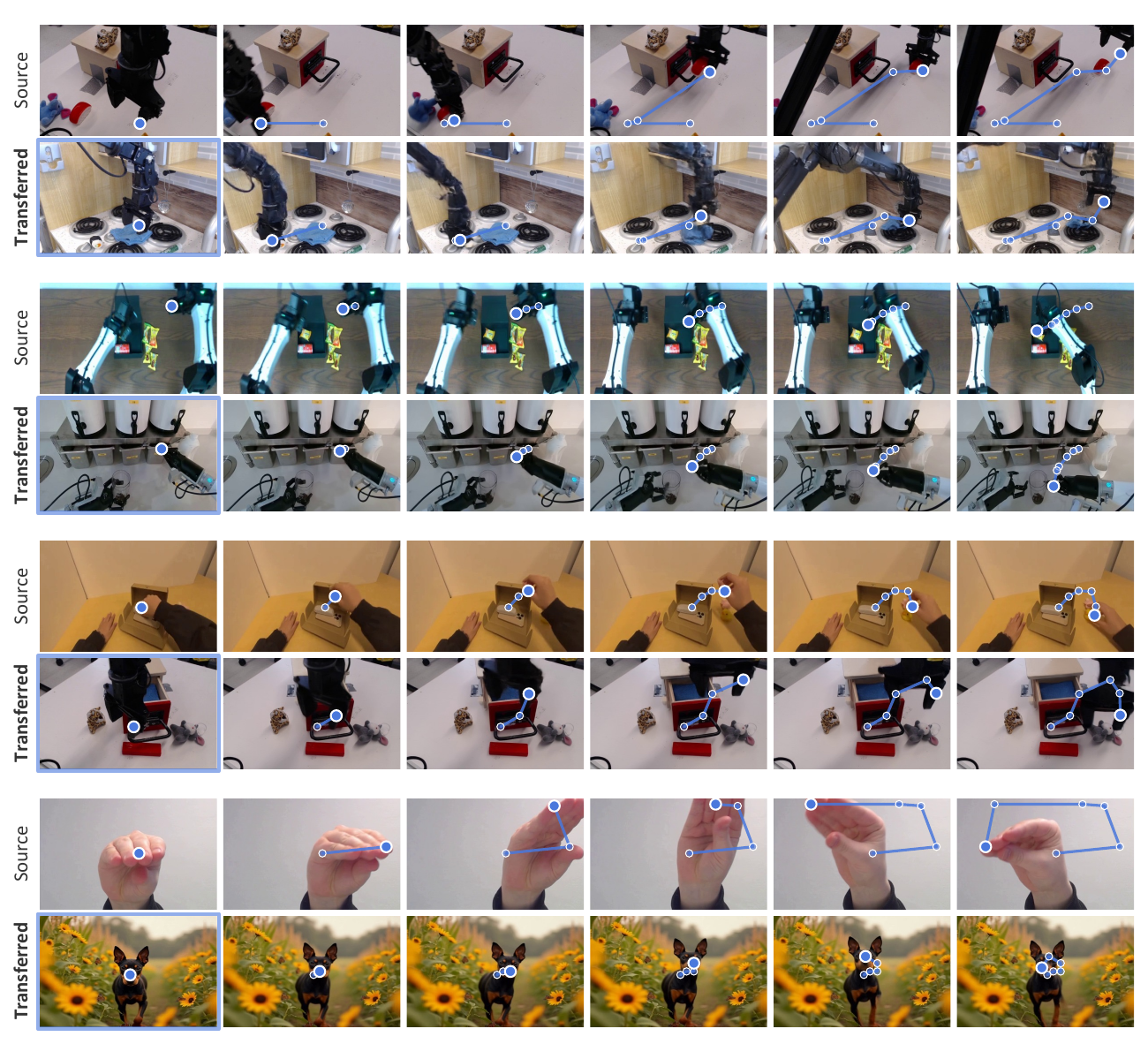}
    \vspace{-18pt}
    \caption{
    \textbf{Qualitative Any-to-Any video transfer.} Each example shows a source motion and the corresponding transferred video generated from a target-domain initial observation. These results demonstrate that the learned latent action transfers motion across substantial changes in embodiment and scene appearance. Blue dots indicate the end-effector trajectory throughout the video frames.
    }
    \label{fig:transfer_qualitative}
\end{figure*}

\subsection{Representation Quality}
\label{sec:representation_quality}

We examine how well \ours{}'s latent actions capture motion on LARYBench~\citep{nie2026lary}. Regression asks whether the representation retains enough fine-grained motion to recover end-effector trajectories, and classification whether it carries enough semantic information to distinguish different motions. Table~\ref{tab:lary_regression_classification} compares \ours{} with latent action models that encode either frame pairs or video clips. \ours{} achieves the best performance across all benchmarks. Compared with the strongest baseline, DiLA, \ours{} reduces regression MSE from 0.38 to 0.17 and increases classification accuracy from 62.25\% to 69.67\%. These results support leveraging video models as a source of action-relevant representations: reading latent actions directly from their pretrained priors yields representations that capture spatial, temporal, and semantic structure across embodiments. FAE-32, our 32-channel compressed representation (Sec.~\ref{sec:latentwam}), also outperforms all baselines, indicating that even the compact codes retain substantial motion and semantic information. Further evaluation details are provided in Appendix~\ref{app:lary_protocol}.

\subsection{Video Transfer}
\label{sec:video_transfer}

We evaluate whether \ours{} learns universal latent actions by transferring source motion across scenes and embodiments while preserving the appearance specified by a target initial frame. We examine any-to-any video transfer and quantitatively evaluate zero-shot human-to-robot transfer. Zero-shot evaluation details and additional transfer examples are provided in Appendices~\ref{app:video_transfer_evaluation} and~\ref{app:video_transfer_qualitative}.

\paragrapht{Any-to-any video transfer.}
Figure~\ref{fig:transfer_qualitative} shows motion transfer across environments for the same robot, between different robot embodiments, and from humans to robots, with a DAVIS example extending beyond manipulation. \ours{} reproduces the source motion while preserving the target scene and embodiment. These results support the universality of the learned latent actions across embodiments and environments, extending to general-domain videos.

\begin{figure*}[t]
    \centering
    \includegraphics[width=\textwidth]
    {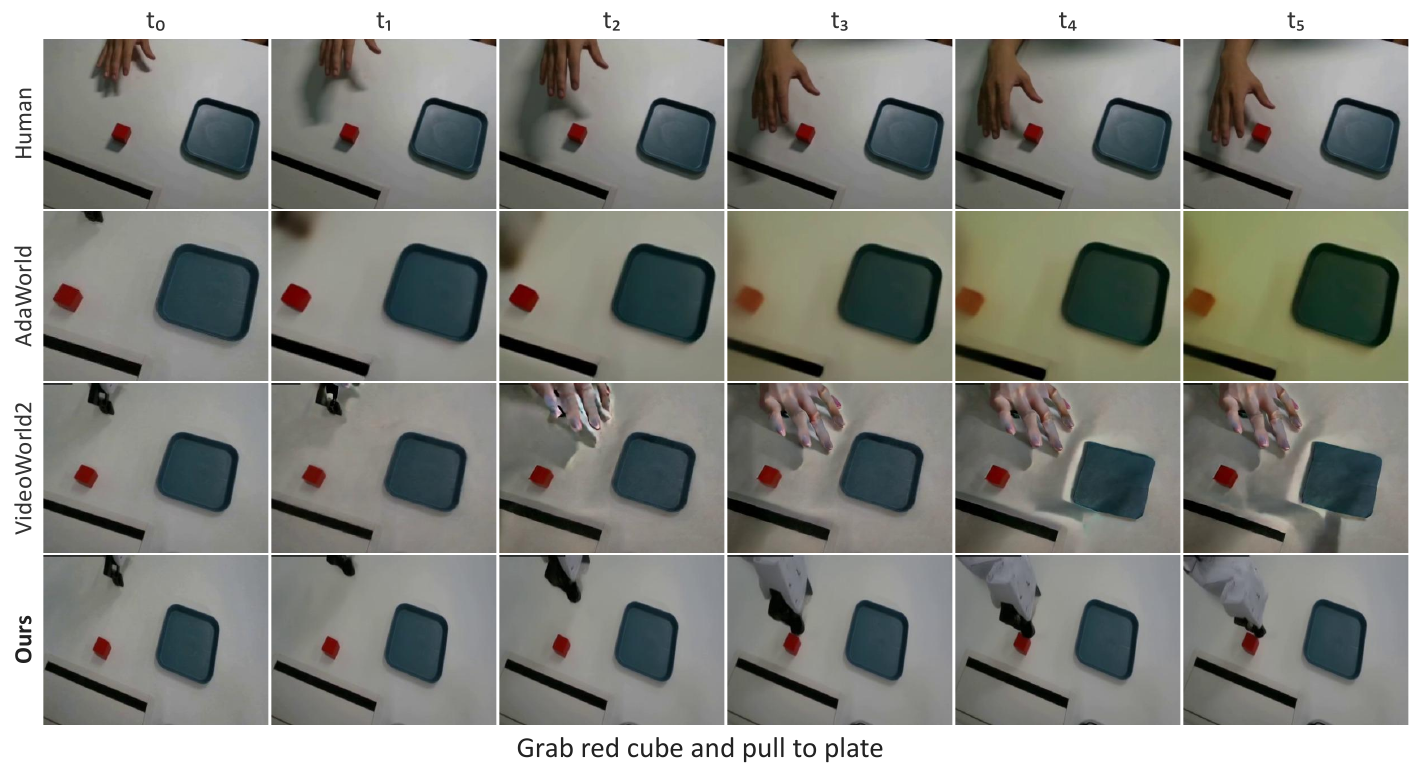}
    \vspace{-1.5em}
    \caption{
    \textbf{Zero-shot human-to-robot video transfer on the H\&R benchmark.}
    DreamX preserves the source motion and target robot appearance, while the baselines exhibit visual degradation or human-hand leakage.
    }
    \label{fig:hr_qualitative}
\end{figure*}

\paragrapht{Zero-shot human-to-robot transfer.}
\newsavebox{\videotransfertablebox}
\begin{lrbox}{\videotransfertablebox}
    \scriptsize
    \setlength{\tabcolsep}{1.5pt}
    \begin{tabular}{lcccc}
        \toprule
        Method & FVD$\downarrow$ & LPIPS$\downarrow$ & PSNR$\uparrow$ & SSIM$\uparrow$ \\
        \midrule
        AdaWorld & 170.19 & 0.344 & 22.95 & 0.846 \\
        VideoWorld2 & 241.92 & 0.299 & 22.53 & 0.881 \\
        \textbf{Ours} & \textbf{57.44} & \textbf{0.179} &
        \textbf{26.52} & \textbf{0.917} \\
        \bottomrule
    \end{tabular}
\end{lrbox}
\begin{wraptable}{r}{\wd\videotransfertablebox}
    \vspace{-14pt}
    \centering
    \captionsetup{width=\linewidth}
    \caption{\textbf{Zero-shot human-to-robot video transfer on H\&R.}
    DreamX outperforms baselines across all metrics
    without paired human--robot training data.}
    \vspace{-0.8em}
    \label{tab:video_transfer}
    \usebox{\videotransfertablebox}
    \vspace{-20pt}
\end{wraptable}

We evaluate zero-shot human-to-robot transfer on HumanAndRobot~\citep{xie2026human2robot}, comparing with AdaWorld~\citep{gao2025adaworld} and VideoWorld~2~\citep{videoworld2}, both of which use diffusion-based video decoders. Paired robot recordings serve only as evaluation references; \textbf{no paired human--robot data is used for training}. \ours{} outperforms both baselines on every metric in Table~\ref{tab:video_transfer} and follows the source human motion while preserving the target robot and scene (Figure~\ref{fig:hr_qualitative}). In contrast, AdaWorld exhibits visual degradation in this out-of-distribution setting, while VideoWorld~2 occasionally introduces a human hand despite its appearance-disentangling design. These results suggest that \ours{} better separates transferable motion from appearance, enabling more consistent human-to-robot transfer.

\subsection[Action Transfer]{Action Transfer}
\label{sec:action_transfer}
\begin{figure*}[t]
    \centering
    \includegraphics[width=\textwidth]{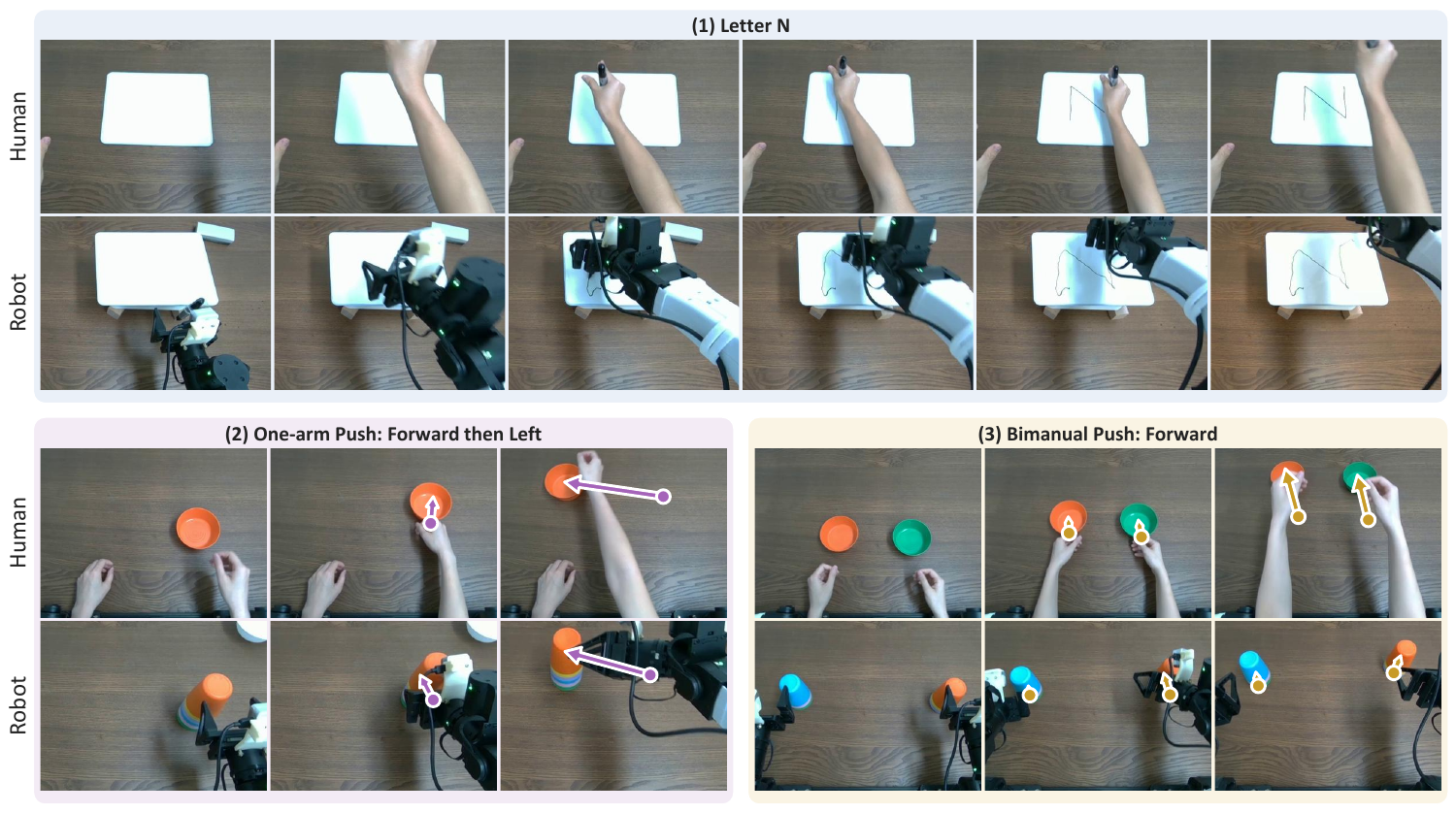}
    \vspace{-15pt}
    \caption{
    \textbf{Qualitative human-to-robot action transfer.} Given a human demonstration, \ours{} extracts a latent action and decodes it into low-level robot actions without paired human--robot training data. We show diverse transferred
    behaviors spanning writing, single-arm pushing, and bimanual manipulation.
    Across substantial embodiment differences, the robot reproduces the motion
    structure of the human demonstration.
    }
    \label{fig:action_transfer_qualitative}
\end{figure*}

We evaluate action transfer to examine whether our latent actions form a universal representation space that connects similar motions across embodiments. Following Sec.~\ref{sec:transfer}, we extract latent actions from a human video and decode them into low-level actions for the target robot. Execution is \emph{open-loop without visual observations}: only the latent actions specify the motion, while proprioception provides the robot's initial state. Notably, this transfer is learned \textbf{without paired human--robot training data}, so the two embodiments are connected through the shared latent action space. Figure~\ref{fig:action_transfer_qualitative} shows that the robot reproduces the stroke structure of the letter ``N,'' preserving the demonstrated motion beyond a coarse task outcome. In single-arm pushing, it follows the forward-then-left sequence, retaining the direction change and temporal order. Bimanual pushing further shows corresponding movements of both arms, suggesting that the shared latent representation retains arm-specific motion and transfers it to the corresponding robot arms. Evaluation protocols and additional examples are provided in Appendices~\ref{app:action_transfer_evaluation} and~\ref{app:action_transfer_qualitative}, respectively.

\subsection{Policy Evaluation}
\label{sec:policy_eval}
% Subgoal success rate uses one point for approach and one for push completion;
% normalize by two and average equally across the three configurations.
% Match the caption and wrapping area to the table's natural width.
\newsavebox{\policydrawertablebox}
\begin{lrbox}{\policydrawertablebox}
    \scriptsize
    \setlength{\tabcolsep}{2pt}
    \renewcommand{\arraystretch}{1.15}
    \begin{tabular}{@{}lccccc@{}}
        \toprule
        Method & Upper & Lower & Both & Avg.$\uparrow$ & \shortstack{Latency$\downarrow$\\(ms)} \\
        \midrule
        Mimic-Video & 50.0 & 50.0 & 50.0 & 50.0 & 28,226 \\
        \ours{} & 90.0 & 100.0 & 100.0 & 96.7 & 1,474 \\
        \bottomrule
    \end{tabular}
\end{lrbox}
\begin{wraptable}{r}{\wd\policydrawertablebox}
    \vspace{-1.2em}
    \centering
    \captionsetup{width=\linewidth}
    \caption{\textbf{Real-world drawer-closing SR (\%) and inference latency.}
    DreamX achieves higher SR than Mimic-Video with $19.1\times$ faster inference,
    predicting compact latent actions instead of future video.}
    \vspace{-0.8em}
    \label{tab:policy_drawer}
    \usebox{\policydrawertablebox}
    \vspace{-1.2em}
\end{wraptable}

% We evaluate \ours{}'s effectiveness and efficiency in real-world control. Given the instruction \emph{close the drawer}, the robot must close the upper drawer, the lower drawer, or both sequentially. We compare \ours{} with Mimic-Video~\citep{pai2025mimic}, both trained using human videos and robot demonstrations. During rollout, Mimic-Video generates future video, while \ours{} generates compact latent actions alongside low-level actions.
We conduct a real-world drawer-closing study to examine whether \ours{}'s learned latent actions can support downstream policy learning. Given the instruction \emph{close the drawer}, the robot must close the upper drawer, the lower drawer, or both sequentially. We compare \ours{} with Mimic-Video~\citep{pai2025mimic}, both trained using human videos and robot demonstrations. During rollout, Mimic-Video generates future video, while \ours{} generates compact latent actions alongside low-level actions.

% Table~\ref{tab:policy_drawer} shows that \ours{} achieves a subgoal success rate of 96.7\%, compared with 50.0\% for Mimic-Video. While Mimic-Video often reaches the drawer without completing the push, \ours{} reliably completes both stages, including sequential drawer closure. \ours{} also reduces inference latency from 28.23~s to 1.47~s per call---a $19.1\times$ speedup. These results suggest that action-centric representations learned from human and robot videos provide more effective conditioning for low-level control than future-video prediction. See Appendices~\ref{app:policy_evaluation}, \ref{app:policy_quantitative}, and~\ref{app:policy_qualitative} for evaluation details and additional results.
Table~\ref{tab:policy_drawer} reports an average subgoal success rate of 96.7\% for \ours{}, averaging reaching and pushing completion within each configuration and weighting the three configurations equally. Pushing completion rates are 80\%, 100\%, and 100\% for the upper, lower, and both-open configurations, respectively. We also report an inference latency of 1.47~s per policy call for \ours{}. These results demonstrate the feasibility of using the learned latent actions as prediction targets for a robot policy in the evaluated drawer-closing setting. See Appendix~\ref{app:policy_evaluation}, \ref{app:policy_quantitative}, and~\ref{app:policy_qualitative} for additional evaluation details and results.

\begin{wrapfigure}{r}{0.39\columnwidth}
    \centering
    \vspace{-13pt}

    % Table
    \begin{minipage}{\linewidth}
        \centering
        \captionof{table}{\textbf{Encoder supervision ablation.}
        LARYBench regression MSE and classification accuracy.}
        \label{tab:encoder_supervision_ablation}

        \vspace{-7pt}
        \resizebox{\linewidth}{!}{%
            % Content-only table, placed in figure/ablation_analysis_row.tex.
% Original compact typesetting: no forced height or stretched row spacing.
\begingroup
    \scriptsize
    \setlength{\tabcolsep}{3pt}
    \begin{tabular}{lcc}
        \toprule
        Supervision
        & \shortstack{Reg. avg.\\$\downarrow$}
        & \shortstack{Cls. avg.\\$\uparrow$} \\
        \midrule
        RGB only                         & 0.24 & 65.3 \\
        \shortstack[l]{RGB + action} & 0.23 & 64.8 \\
        \bottomrule
    \end{tabular}
\endgroup

        }
    \end{minipage}

    \par\vspace{7pt}

    % Figure
    \begin{minipage}{\linewidth}
        \centering
        \includegraphics[width=\linewidth]{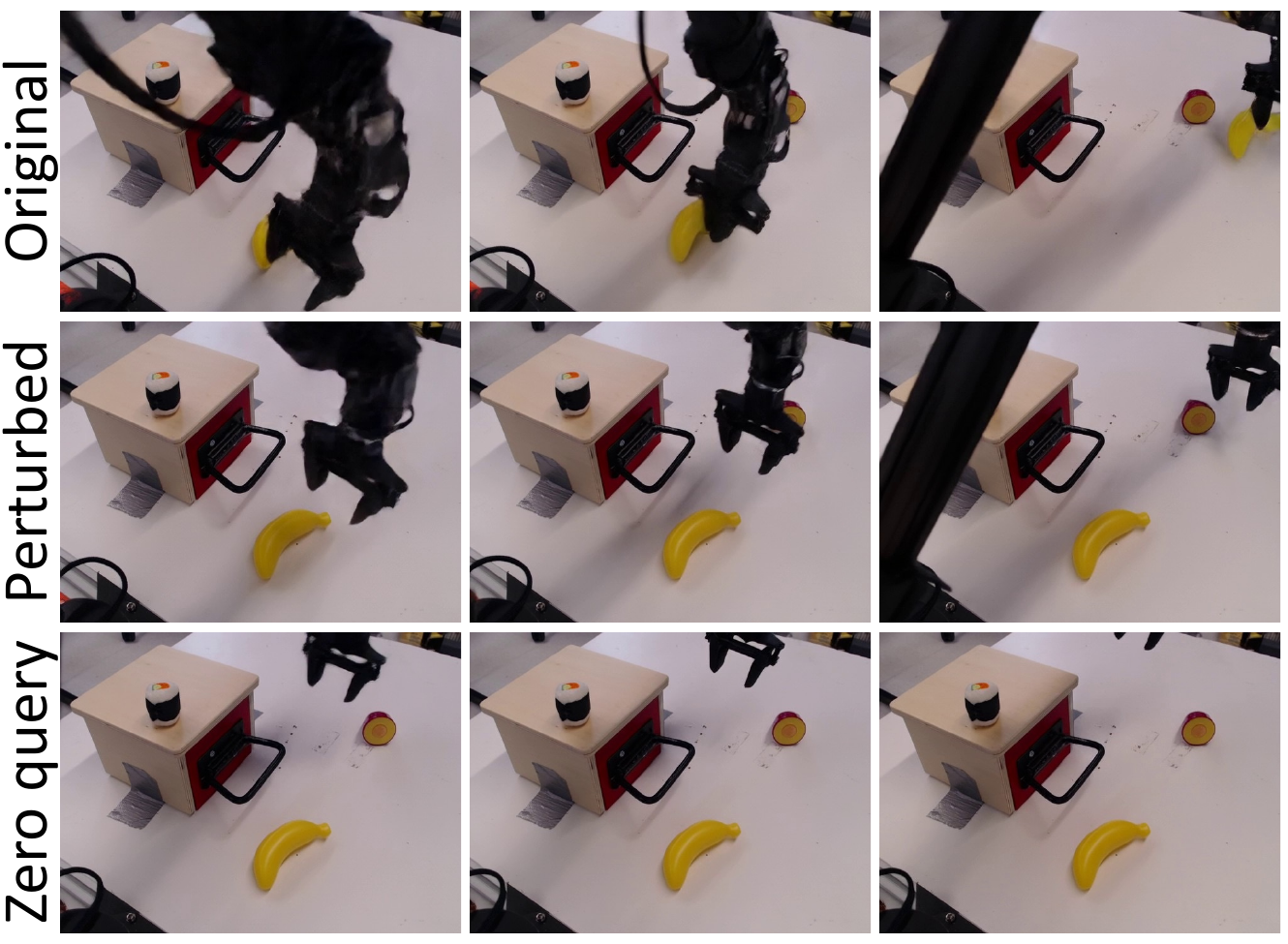}
        \vspace{-15pt}
        \caption{\textbf{Sensitivity analysis.}
        Perturbing latent actions alters motion while preserving
        scene appearance.}
        \vspace{-40pt}
        \label{fig:latent_perturbation}
    \end{minipage}
\end{wrapfigure}
 % above two merged

\subsection{Ablation Study}
\label{sec:ablation}

We examine how auxiliary action supervision affects \ours{}'s latent actions, using the same subsets of MolmoAct2, EgoVerse, and EgoDex to limit computational cost. Table~\ref{tab:encoder_supervision_ablation} shows that action supervision slightly improves regression MSE from $0.24$ to $0.23$, while classification accuracy slightly decreases from $65.3\%$ to $64.8\%$. The comparable performance suggests that \ours{} captures motion and action semantics from video alone, without relying on action labels. 
% Auxiliary action supervision also supports applications such as human-to-robot action transfer, allowing latent actions extracted from human videos to be decoded directly into executable robot actions without further adaptation.

\subsection{Analysis}
\label{sec:analysis}
\paragraph{Sensitivity to latent-action perturbations.}

\begin{figure}[t]
    \centering
    \includegraphics[width=\columnwidth]{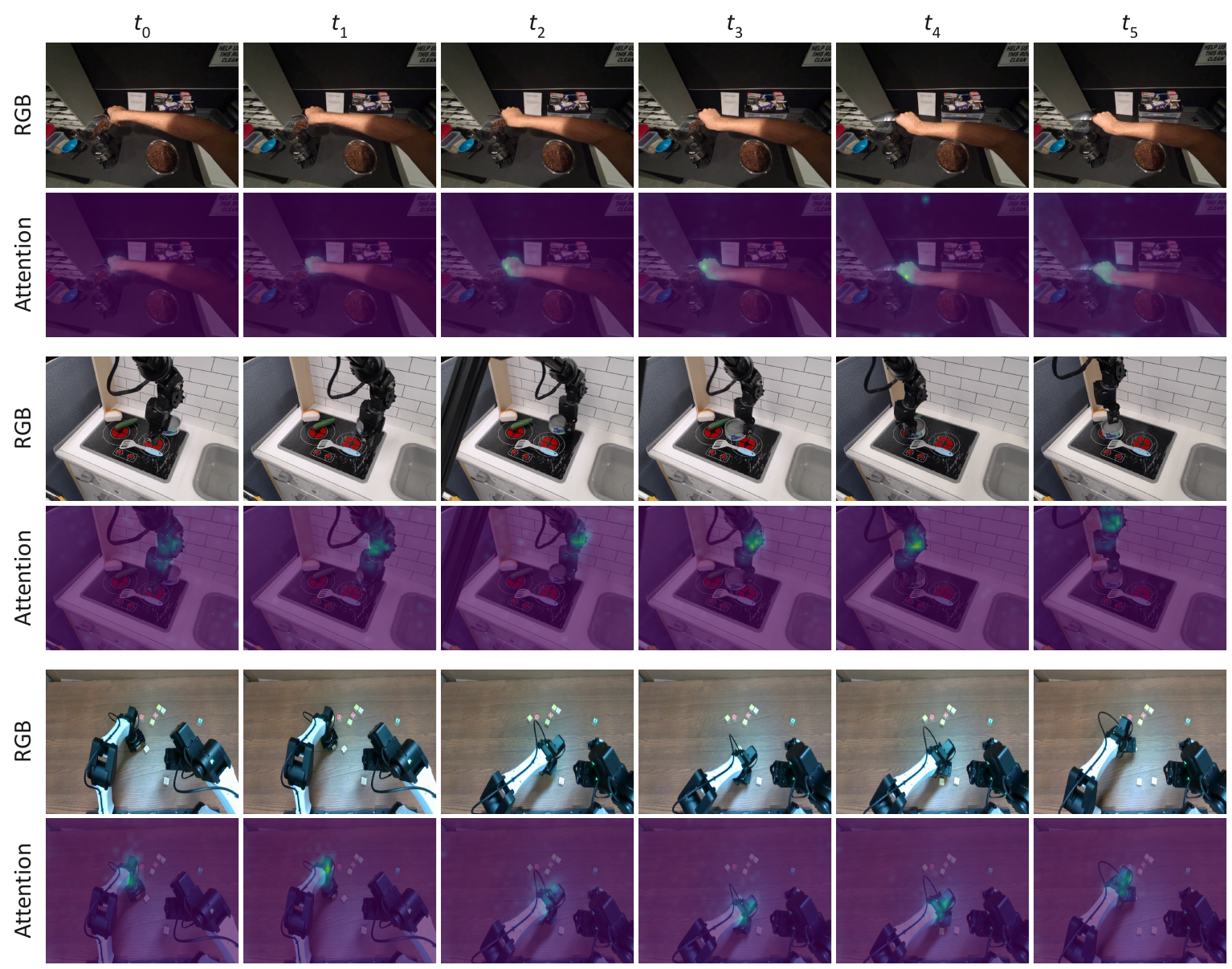}
    \caption{\textbf{Query-to-video attention.}
    We visualize attention maps from latent queries to video tokens during encoding. Across human and robot demonstrations, latent queries attend to dynamic regions, such as moving hands and grippers, capturing the underlying actions in each video.}
    \label{fig:query_attention}
\end{figure}

We examine whether latent actions guide the motion reconstructed by the decoder. Keeping the initial frame and sampling noise fixed, we perturb the latent actions or replace them with zeros. In Figure~\ref{fig:latent_perturbation}, the original latent actions reconstruct the grasping and lifting motion, whereas perturbation disrupts object manipulation and zeroing largely suppresses motion. The surrounding scene remains largely unchanged, supporting the role of latent actions in specifying motion beyond appearance. Quantitative results are provided in Appendix~\ref{app:latent_conditioning}.

\paragrapht{What latent actions attend to.}

To examine where \ours{} reads motion information, Figure~\ref{fig:query_attention} visualizes query-to-video attention during latent-action encoding of human and robot demonstrations. Attention highlights the hands and grippers involved in object interactions, suggesting that the queries read action-relevant regions across embodiments. Additional visualizations are provided in Appendix~\ref{app:query_attention}.

\section{Conclusion}
\label{sec:conclusion}

We present \ours{}, a framework that leverages pretrained video diffusion priors to learn universal latent actions. Its spatiotemporal representations provide action-relevant dynamics, while its generative prior supervises latent-action learning through future-frame reconstruction. \ours{} achieves state-of-the-art performance among latent action models on LARYBench and enables any-to-any video transfer and human-to-robot action transfer without paired training data. Predicting these compact latent actions also supports effective and efficient policy learning from human and robot demonstrations. Together, these results highlight pretrained video models as a foundation for learning reusable action representations across embodiments.

\clearpage

\subsection*{AI use statement}

We used generative AI tools to assist with manuscript drafting and editing for clarity and readability, code editing during implementation, and discussions during the research process. Research ideas, methodological choices, experimental designs, and interpretation of results were formulated, evaluated, and ultimately determined by the authors. All AI-assisted text and code were reviewed by the authors, and code changes were tested for correctness. We take responsibility for the final content of this work, including text, claims, code, and artifacts produced with the aid of generative AI.

\subsection*{Ethics statement}

All datasets used in this work are publicly available, including the human egocentric and robot demonstration datasets listed in Appendix~\ref{app:training_data}. Human videos may contain identifiable people or surroundings, so any redistributed examples should follow the source datasets' usage terms and privacy requirements. Video-transfer outputs are synthetic and should be identified as such. Cross-embodiment transfer and robot policies may fail outside the evaluated settings and require safety validation before deployment.

\subsection*{Reproducibility statement}

Section~\ref{sec:method} describes the model, training objectives, transfer procedure, and policy. Appendix~\ref{app:implementation} documents the data sources and implementation details, while Appendix~\ref{app:evaluation_setup} specifies the benchmark, transfer, and real-world policy evaluation protocols. Supplementary videos illustrate video-transfer and action-transfer results; policy rollouts are provided in Appendix~\ref{app:policy_qualitative}.

% \subsubsection*{Author Contributions}
% If you'd like to, you may include  a section for author contributions as is done
% in many journals. This is optional and at the discretion of the authors.

% \subsubsection*{Acknowledgments}
% Use unnumbered third level headings for the acknowledgments. All
% acknowledgments, including those to funding agencies, go at the end of the paper.

\bibliography{iclr2027_conference}
\bibliographystyle{iclr2027_conference}

\clearpage

\appendix

This appendix provides implementation details, evaluation protocols, additional results and analysis.
\begin{itemize}
    \item[] \textbf{\ref{app:implementation}. IMPLEMENTATION DETAILS}
    \begin{itemize}
        \item[] \ref{app:training_data} Training Data
        \item[] \ref{app:pretraining_details} DreamX Pretraining
        \item[] \ref{app:fae_details} Latent-Action Compression
        \item[] \ref{app:policy_implementation} Policy Training and Inference
    \end{itemize}
    \item[] \textbf{\ref{app:evaluation_setup}. EXPERIMENTAL SETUP AND EVALUATION PROTOCOLS}
    \begin{itemize}
        \item[] \ref{app:real_world_setup} Real-World Robot Setup
        \item[] \ref{app:action_transfer_evaluation} Action Transfer Details
        \item[] \ref{app:policy_protocol} Policy Evaluation Details
        \item[] \ref{app:lary_protocol} LARYBench Evaluation Details
        \item[] \ref{app:video_transfer_evaluation} Zero-Shot Human-to-Robot Transfer Details
    \end{itemize}
    \item[] \textbf{\ref{app:quantitative_results}. ADDITIONAL QUANTITATIVE RESULTS}
    \begin{itemize}
        \item[] \ref{app:policy_quantitative} Policy Evaluation
        \item[] \ref{app:calvin_ablation} Ablation Study
        \item[] \ref{app:latent_analysis} Representation Analysis
    \end{itemize}
    \item[] \textbf{\ref{app:qualitative_results}. ADDITIONAL QUALITATIVE RESULTS AND ANALYSIS}
    \begin{itemize}
        \item[] \ref{app:video_transfer_qualitative} Any-to-Any Video Transfer
        \item[] \ref{app:action_transfer_qualitative} Human-to-Robot Action Transfer
        \item[] \ref{app:policy_qualitative} Policy Rollouts
        \item[] \ref{app:query_attention} Query-to-Video Attention
    \end{itemize}
    % \item[] \textbf{\ref{app:limitations}. LIMITATIONS AND FUTURE WORK}
\end{itemize}
Video transfer, action transfer videos are provided in the supplementary materials.
\vspace{0.5cm}

\section{Implementation Details}
\label{app:implementation}

\subsection{Training Data}
\label{app:training_data}

\begin{figure}[htbp]
    \centering
    \includegraphics[width=0.72\linewidth]{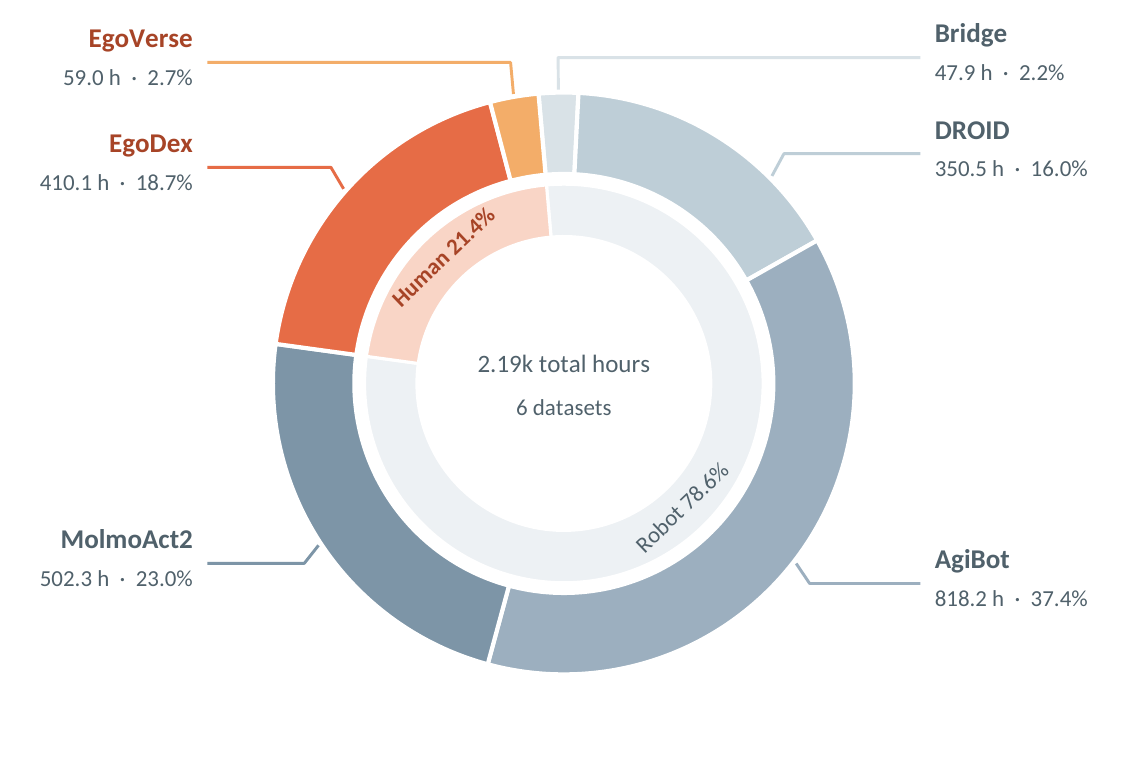}
    \caption{\textbf{Pretraining data composition.} DreamX learns from six datasets spanning human and robot demonstrations. Sector areas represent video duration, with human datasets highlighted in warm colors.}
    \label{fig:pretraining_data_composition}
\end{figure}

% We pretrain on a mixture of robot and human video. Robot data comes from
% MolmoAct-BimanualYAM~\citep{fang2026molmoact2}, Bridge~\citep{walke2023bridgedata},
% DROID~\citep{khazatsky2024droid}, and AgiBot~\citep{bu2025agibot}, and human
% data from EgoDex~\citep{hoque2026egodex} and EgoVerse~\citep{punamiya2026egoverse}.
% Figure~\ref{fig:pretraining_data_composition} summarizes the mixture, comprising
% 2.19k hours of video, including 469 hours of human demonstrations.
We pretrain \ours{} on 2.19k hours of human and robot video spanning six datasets
(Figure~\ref{fig:pretraining_data_composition}). Human demonstrations contribute
469 hours from EgoDex~\citep{hoque2026egodex} and EgoVerse~\citep{punamiya2026egoverse},
complementing robot demonstrations from Bridge~\citep{walke2023bridgedata},
DROID~\citep{khazatsky2024droid}, AgiBot~\citep{bu2025agibot}, and
MolmoAct-BimanualYAM~\citep{fang2026molmoact2}.

\subsection{DreamX Pretraining}
\label{app:pretraining_details}

% We build on Cosmos-Predict2.5 2B at $480 \times 640$. We train with variable
% length clips of $T \in \{5,9,13,17,21,25,29\}$ pixel frames, corresponding to
% $N \in \{2,\ldots,8\}$ VAE latent frames. Each query-token group consists of
% $K=4$ tokens and corresponds to one VAE latent frame. The action loss is
% weighted with $\lambda=1$ for clips carrying action labels and $\lambda=0$ otherwise.
We initialize \ours{} from Cosmos-Predict2.5 2B and train at $480 \times 640$
resolution for 78,000 steps. Training clips contain 5--29 frames in increments
of four, corresponding to 2--8 VAE latent frames. We assign four query tokens
to each latent frame, yielding up to 32 query tokens per clip. The encoder and
decoder use separate rank-16 LoRA modules, while the pretrained backbone weights
remain frozen. Learning rates are $10^{-4}$ for the LoRA parameters and query
embeddings and $5 \times 10^{-5}$ for the remaining trainable parameters, with
500 warm-up steps. Video and action losses receive equal weights for
action-labeled clips; clips without action labels use only the video loss.

\subsection{Latent-Action Compression}
\label{app:fae_details}

% We train a variational bottleneck on latent actions from the frozen \ours{}
% encoder. A single self-attention block encodes each 2,048-dimensional token into
% a 32-dimensional code, and a six-layer transformer decoder reconstructs the
% original features. The token layout is preserved, mapping a $32 \times 2{,}048$
% representation to $32 \times 32$ compact codes. Training uses
% feature-reconstruction MSE in a fixed normalized feature space and a KL
% regularizer with weight $\beta=0.01$. We train for 50,000 steps with a learning
% rate of $10^{-4}$ and 500 warm-up steps. After training, we use the posterior
% mean to obtain deterministic compact codes for downstream learning and evaluation.

We train a variational bottleneck on latent actions $Q$ extracted by the frozen \ours{} encoder. Following~\citep{fae}, the bottleneck compresses each token from backbone width $D$ to $d \ll D$ while preserving the token layout.

\paragrapht{Formulation.}
The encoder $f_\psi$ parameterizes a factorized Gaussian posterior $q_\psi(Z \mid Q)$ through
\[
    \big(\mu_\psi(Q),\,\log \sigma_\psi^2(Q)\big)
    = f_\psi(Q).
\]
During bottleneck training, we sample compact codes using
\begin{equation}
    Z = \mu_\psi(Q) + \sigma_\psi(Q) \odot \epsilon,
    \qquad
    \epsilon \sim \mathcal{N}(\mathbf{0}, \mathbf{I}),
    \label{eq:fae-encode}
\end{equation}
and reconstruct the original latent actions as $\hat{Q}=g_\psi(Z)$.

Only the bottleneck encoder $f_\psi$ and decoder $g_\psi$ are optimized, using
\begin{equation}
    \mathcal{L}_{\mathrm{FAE}}
    = \big\lVert \hat{Q} - Q \big\rVert_2^2
    + \beta\,\mathrm{KL}\big(
        q_\psi(Z \mid Q)
        \,\Vert\,
        \mathcal{N}(\mathbf{0}, \mathbf{I})
    \big).
    \label{eq:fae-objective}
\end{equation}
The reconstruction term encourages the compact codes to retain the information in $Q$, while the KL term regularizes their posterior toward a standard Gaussian. We compute the reconstruction loss in a fixed normalized feature space.

\paragrapht{Architecture and Training.}
The bottleneck encoder uses a single self-attention block to produce a 32-dimensional code for each 2,048-dimensional latent-action token. A six-layer transformer decoder reconstructs the original features. The token layout is preserved: for a 32-token input, the bottleneck maps a $32 \times 2{,}048$ representation to $32 \times 32$ compact codes.

We train the bottleneck for 50,000 steps with a learning rate of $10^{-4}$, 500 warm-up steps, and a KL weight of $\beta=0.01$. After training, we freeze the bottleneck and use the posterior mean,
\[
    Z = \mu_\psi(Q),
\]
to obtain deterministic compact codes for downstream policy learning and evaluation.

\subsection{Policy Training and Inference}
\label{app:policy_implementation}

To convert the pretrained video model into a latent-action generator, we train
it to predict compact latent-action codes conditioned on the current observation
and language instruction. We use approximately one-third of the \ours{}
pretraining data and optimize rank-128 LoRA modules in self-attention,
cross-attention, and feed-forward layers, together with trainable input and
output projections. The generator is trained for 100,000 steps on two NVIDIA
H200 GPUs with a global batch size of 60, a learning rate of $10^{-4}$, and
1,000 warm-up steps. The Action DiT is not separately pretrained; it is
introduced during task-specific policy learning and trained jointly with the
generator. Task-specific training settings are provided in
Appendix~\ref{app:policy_protocol}.

\section{Experimental Setup and Evaluation Protocols}
\label{app:evaluation_setup}

\begin{wrapfigure}{r}{0.35\columnwidth}
  \centering
  \vspace{-40pt}
  \includegraphics[
  width=0.9\linewidth,
  trim={0pt 0pt 0pt 20pt},
  clip
]{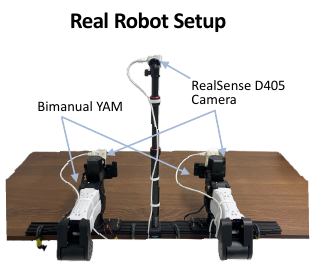}
  \vspace{-10pt}
  \caption{\textbf{BimanualYAM setup.}}
  \vspace{-30pt}
  \label{fig:suppl_robot_setup}
\end{wrapfigure}

\subsection{Real-World Robot Setup}
\label{app:real_world_setup}
\label{app:robot_hardware}

We conduct both action-transfer and policy experiments on a bimanual YAM robot.
The action space consists of 14-dimensional absolute commands, comprising six
joint positions and one gripper command per arm. We leverage one Intel RealSense D405 camera for observation, which is mounted on top view.

\subsection{Action Transfer Details}
\label{app:action_transfer}
\label{app:action_transfer_evaluation}

We extract latent actions from human demonstrations using the frozen \ours{}
encoder and decode them into robot commands. The resulting trajectory is
executed \emph{open-loop, without visual feedback}, conditioned on the human
latent actions and the robot's initial proprioceptive state.

We evaluate three types of motion transfer. \textbf{Writing} tests whether
latent actions preserve fine-grained spatial trajectories and stroke order;
a marker attached to the right gripper reproduces the demonstrated shape.
\textbf{Single-arm pushing} tests whether the representation captures changes
in motion direction and their temporal order, requiring the robot to follow
a forward-then-left sequence. \textbf{Bimanual manipulation} tests whether
latent actions retain arm-specific motion and transfer it to the corresponding
robot arms. Additional examples are provided in
Section~\ref{app:action_transfer_qualitative}.

\subsection{Policy Evaluation Details}
\label{app:policy_evaluation}
\label{app:policy_protocol}

\begin{figure}[htbp]
    \centering
    \begin{minipage}[t]{0.325\linewidth}
        \centering
        \includegraphics[width=\linewidth]{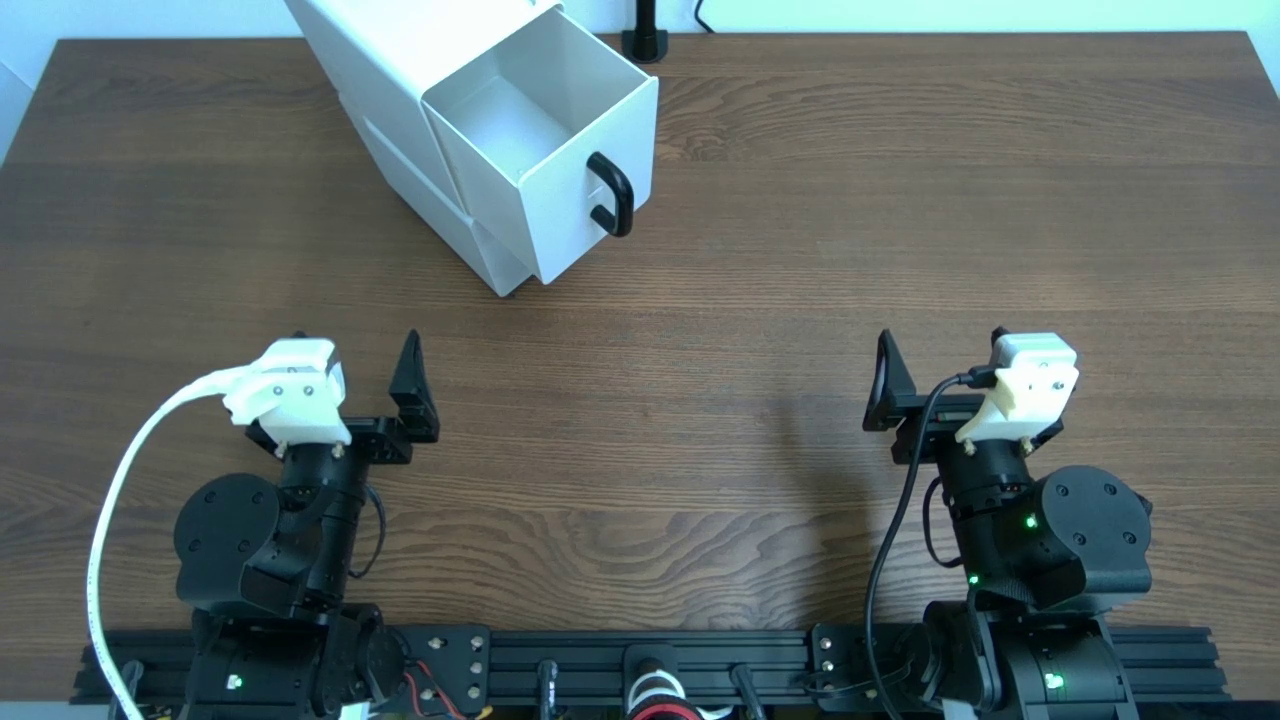}
        \par\smallskip
        {\small (a) Upper drawer open}
    \end{minipage}\hfill
    \begin{minipage}[t]{0.325\linewidth}
        \centering
        \includegraphics[width=\linewidth]{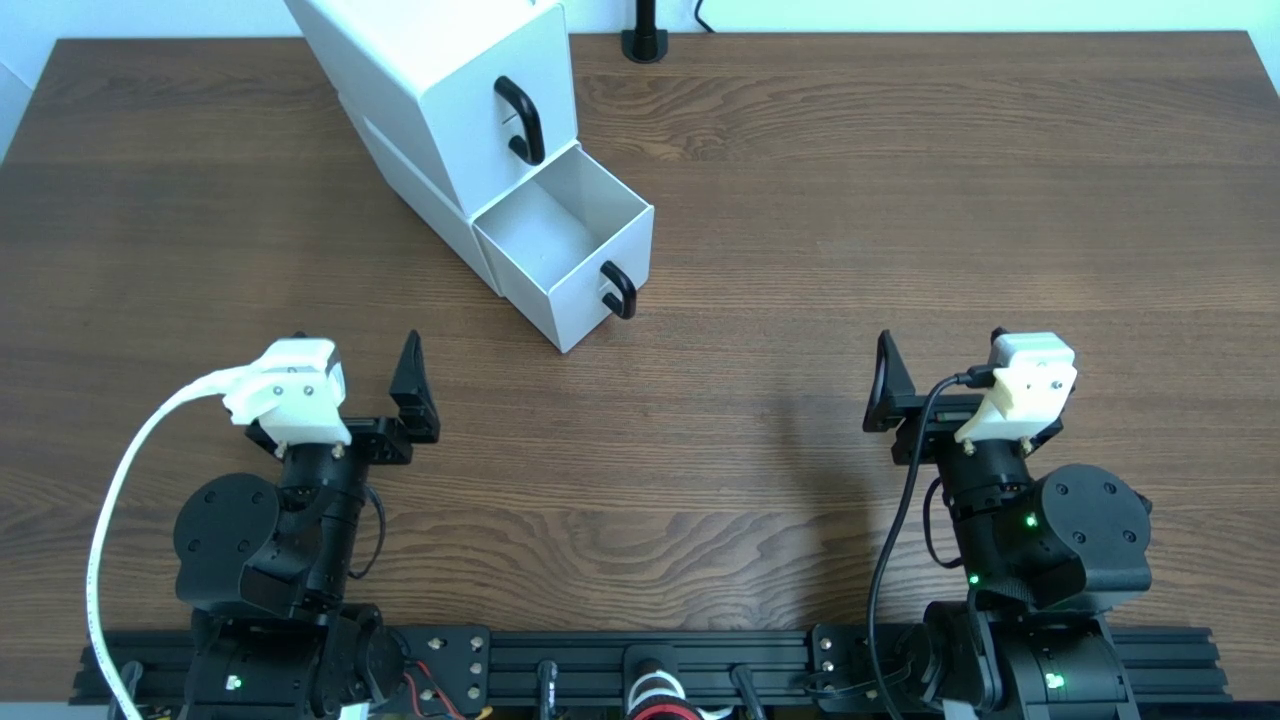}
        \par\smallskip
        {\small (b) Lower drawer open}
    \end{minipage}\hfill
    \begin{minipage}[t]{0.325\linewidth}
        \centering
        \includegraphics[width=\linewidth]{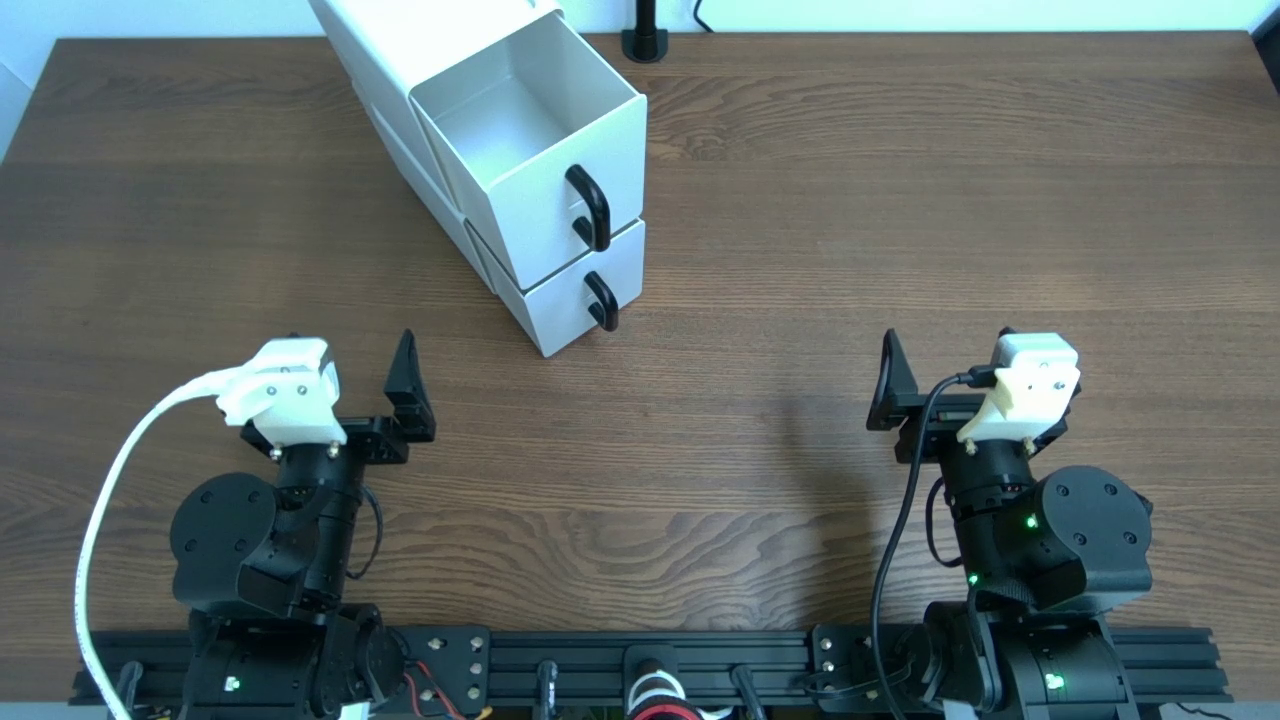}
        \par\smallskip
        {\small (c) Both drawers open}
    \end{minipage}
    \caption{\textbf{Real-world drawer-closing setup.} Initial observations for the three task configurations, all conditioned on \emph{close the drawer}. When both drawers are open, the robot must close the upper drawer before the lower drawer.}
    \label{fig:policy_initial_states}
\end{figure}

\paragraph{Task and observations.}
We evaluate three drawer-closing configurations: only the upper drawer open,
only the lower drawer open, and both drawers open, as illustrated in
Figure~\ref{fig:policy_initial_states}. All configurations use the instruction
\emph{close the drawer}. When both drawers are open, the robot must close the
upper drawer before the lower drawer. A fixed top-view RGB camera observes
the workspace.

\paragraph{Training.}
We compare \ours{} with Mimic-Video~\citep{pai2025mimic}, using human videos and
robot demonstrations for task-specific learning. \ours{} starts from the
pretrained latent-action generator, while Mimic-Video starts from a video
backbone pretrained on MolmoAct. Each method then incorporates an action head
for task-specific learning, and both are trained for 30,000 steps. For \ours{},
human videos supervise latent-action prediction, while robot demonstrations
additionally supervise low-level actions. During rollout, \ours{} generates
compact latent actions alongside low-level actions, whereas Mimic-Video
generates future video.

\paragraph{Evaluation.}
We evaluate reaching and pushing as two subgoals: reaching measures whether
the robot approaches the drawer for interaction, and pushing measures whether
it completes the required closure. The reported subgoal success rate averages
the two completion rates within each configuration, and the overall average
weights the three configurations equally. We additionally report inference
latency per policy call. Separate reaching and pushing results are provided in
Section~\ref{app:policy_quantitative}.
We collected 45 trajectories to train the policy and evaluated its success rate over 10 trials in each of three scenarios.

\subsection{LARYBench Evaluation Details}
\label{app:lary}
\label{app:lary_protocol}

\paragraph{Evaluation protocol.}
We freeze all representation models and train a common attentive probe for
action regression and classification. Each input token is projected to 768
dimensions through a linear layer, layer normalization, and GELU. For
representations spanning multiple transitions, learned temporal embeddings
identify the transition positions before the temporal and token dimensions
are flattened. A learned-query attentive pooler aggregates the tokens into a
single feature for the regression or classification head.

\paragrapht{Feature extraction.}
\label{app:lary_caching}
\label{app:lary_baselines}

The regression benchmark originally provides only two frames per sample.
Using the accompanying metadata, we recover the corresponding video sequences
from the source datasets, including the intervening frames, for video-based models.

We extract each model's native representation without fine-tuning and preserve
its temporal and token structure. Pair-based models encode consecutive frame
pairs, while video-based models encode clips using their supported temporal
windows. Features are precomputed for probe training.
\subsection{Zero-Shot Human-to-Robot Transfer Details}
\label{app:video_transfer}
\label{app:video_transfer_evaluation}
We evaluate on 1,008 clips spanning 21 tasks from
HumanAndRobot~\citep{xie2026human2robot}, selecting 48 clips per task.
Each method receives a human source video and the initial frame of its paired
robot recording. The remaining robot frames serve only as evaluation references;
\emph{no paired human--robot data is used for training}. We compare \ours{} with
AdaWorld~\citep{gao2025adaworld} and VideoWorld~2~\citep{videoworld2} using their
native generation horizons.

For a common evaluation horizon, we retain the first 20 frames of each output
and exclude the conditioning frame, yielding 19 generated frames per clip.
We report FVD using a Kinetics-400-pretrained I3D network, VGG-based LPIPS,
PSNR, and SSIM against the corresponding robot reference videos.

\section{Additional Quantitative Results}
\label{app:quantitative_results}

\subsection{Policy Evaluation}
\label{app:policy_quantitative}

\paragraph{Subgoal completion}

Table~\ref{tab:policy_subgoal_breakdown} reports reaching and pushing
completion separately. Averaging each pair gives the per-configuration SR
in Table~\ref{tab:policy_drawer}: 50.0\% for Mimic-Video in all configurations,
and 90.0\%, 100.0\%, and 100.0\% for \ours{}.

\begin{table}[h]
    \centering
    \caption{\textbf{Reaching and pushing completion rates (\%) for real-world drawer
    closing.} Each subgoal contributes equally to the SR reported in the main table.}
    \vspace{-7pt}
    \label{tab:policy_subgoal_breakdown}
    \small
    \begin{tabular}{lcccccc}
        \toprule
        & \multicolumn{2}{c}{Upper} & \multicolumn{2}{c}{Lower} & \multicolumn{2}{c}{Both} \\
        \cmidrule(lr){2-3}\cmidrule(lr){4-5}\cmidrule(lr){6-7}
        Method & Reach & Push & Reach & Push & Reach & Push \\
        \midrule
        Mimic-Video & 80 & 20 & 100 & 0 & 100 & 0 \\
        \ours{} & 100 & 80 & 100 & 100 & 100 & 100 \\
        \bottomrule
    \end{tabular}
\end{table}

\paragraph{Policy learning from human demonstrations}
\label{app:human_demo_substitution}

We additionally train a policy using \emph{only human videos for the upper-only
drawer-closing configuration}, while retaining robot demonstrations for the
lower-only and both-open configurations. Human videos supervise latent-action
prediction without low-level action labels. As shown in
Table~\ref{tab:policy_human_substitution}, the policy achieves a 100\% subgoal success rate
in upper drawer task even if trained only with human dataset.

\begin{table}[h]
    \centering
    \caption{\textbf{Policy learning with human-only demonstrations of the
    upper-only configuration.} Task columns report approach/push completion
    rates (\%); SR denotes subgoal success rate (\%), averaged equally over
    reaching and pushing and across the three configurations.}
    \vspace{-7pt}
    \label{tab:policy_human_substitution}
    \small
    \begin{tabular}{lcccc}
        \toprule
        Method & Upper & Lower & Both & SR$\uparrow$ \\
        \midrule
        \ours{} & 100/100 & 100/100 & 100/100 & 100 \\
        \bottomrule
    \end{tabular}
\end{table}
% Migrated from the former main-table Ours (R+H) rows:
% unfilled row: --/--, --/--, --/--, SR --, latency --.
% filled row (formerly annotated "Robot Part"): 100/100 in all configurations,
% SR 100, latency --. No additional measurements are inferred here.

\clearpage

\subsection{Ablation Study}
\label{app:calvin_ablation}

\paragraph{Latent-action conditioning on CALVIN.}
We compare the \ours{} policy with and without allowing the action head to
attend to latent-action features. In the variant without latent actions, we
block the action head's attention to these features.
Table~\ref{tab:calvin_latent_ablation}
reports the percentage of evaluation sequences completing at least $k$
consecutive tasks, for $k=1,\ldots,5$, and the average completed sequence length.
Conditioning on latent actions improves success rates at all five sequence
lengths and increases the average length from 2.378 to 2.643. These results
suggest that latent-action information provides useful guidance for robot
action prediction in the \ours{} policy.

\begin{table}[htbp]
    \centering
    \captionsetup{width=0.82\linewidth}
    \caption{\textbf{CALVIN policy ablation with and without latent-action conditioning.}
    Columns $\geq k$ report success rates (\%) for completing at least $k$
    consecutive tasks; Avg. Len. is the average completed sequence length.}
    \label{tab:calvin_latent_ablation}
    \small
    \setlength{\tabcolsep}{4pt}
    \renewcommand{\arraystretch}{1.15}
    \begin{tabular}{@{}lcccccc@{}}
        \toprule
        Policy & $\geq 1$ & $\geq 2$ & $\geq 3$ & $\geq 4$ & $\geq 5$
        & Avg. Len.$\uparrow$ \\
        \midrule
        \ours{} policy w/o latent actions
        & 83.4 & 61.7 & 42.5 & 29.8 & 20.4 & 2.378 \\
        \ours{} policy
        & \textbf{85.7} & \textbf{67.0} & \textbf{49.5}
        & \textbf{36.4} & \textbf{25.7} & \textbf{2.643} \\
        \bottomrule
    \end{tabular}
\end{table}

\subsection{Representation Analysis}
\label{app:latent_analysis}

\paragraph{Motion structure across embodiments}
\label{app:latent_clustering}

\begin{wrapfigure}{r}{0.57\linewidth}
    \centering
    \vspace{-10pt}
    \includegraphics[width=\linewidth]{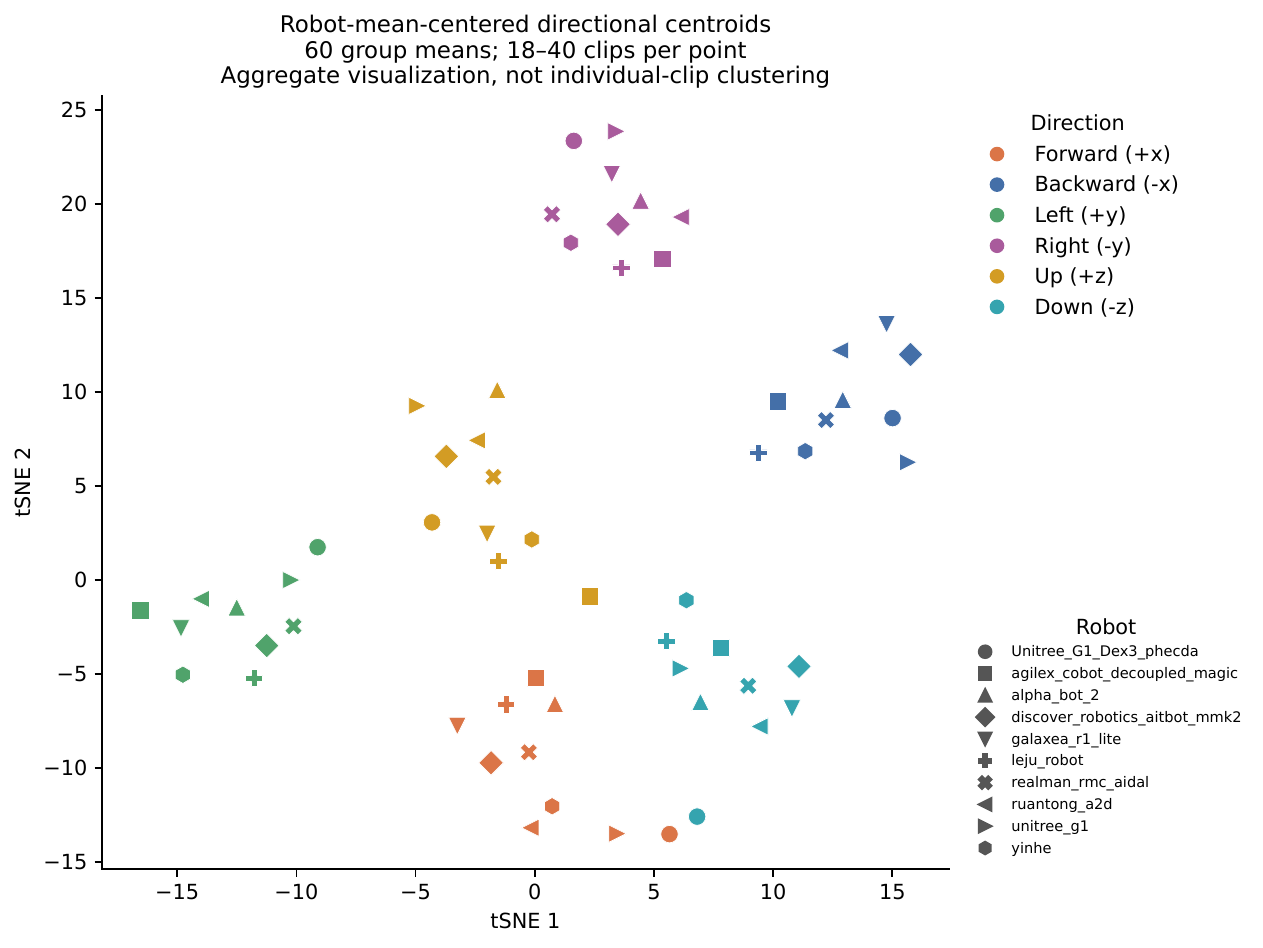}
    \caption{\textbf{Directional structure across robot embodiments.}
    Colors denote motion directions and marker shapes denote robot types.}
    \vspace{-20pt}
    \label{fig:latent_direction_centroids}
\end{wrapfigure}

We examine whether latent actions exhibit common directional structure
across robot embodiments using 2,362 RoboCOIN clips spanning ten robot
types and six end-effector translation directions.

Figure~\ref{fig:latent_direction_centroids} visualizes 60
direction-conditioned latent-action centroids using t-SNE.
Centroids for the same direction exhibit similar organization across
robot types.
% In the original 2,048-dimensional centered feature space, mean
% cross-robot cosine similarity is 0.525 for matching directions
% and $-0.103$ for different directions.
This aggregate analysis supports shared directional structure
across embodiments.

\paragraph{Effect of latent-action conditioning}
\label{app:latent_conditioning}

We quantify how changes to latent actions affect decoded motion and appearance.
For latent actions $Q\in\mathbb{R}^{N_q\times D}$, we perturb the decoder's
conditioning as
\begin{equation}
    \widetilde{Q}_{i,d}=Q_{i,d}+\alpha s_d\epsilon_{i,d},
    \qquad \epsilon_{i,d}\sim\mathcal{N}(0,1),
\end{equation}
where $s_d$ is the channel-wise standard deviation estimated from training
features. We evaluate $\alpha\in\{0,0.1,0.3,1.0\}$ and a zero-query control
on eight selected high-motion clips, with two decoder seeds per clip. Within
each comparison, we fix the initial frame, decoder noise, denoising schedule,
and perturbation draw, rescaling only the perturbation magnitude. The stronger
qualitative perturbation in Figure~\ref{fig:latent_perturbation} uses $\alpha=2$.

Table~\ref{tab:latent_perturbation_metrics} reports mean changes relative to the
unperturbed reconstruction, and Figure~\ref{fig:latent_perturbation_boxplots}
shows their distributions across clips and decoder seeds. We compute Farneb\"ack optical flow at
$160\times120$ and measure mean flow-vector deviation and motion magnitude
relative to the unperturbed output. RGB differences use intensities in $[0,1]$
and exclude the conditioning frame. A fixed static-region mask selects pixels
below the 40th percentile of temporal RGB variation in the reference video,
followed by $5\times5$ erosion.

Increasing perturbation strength produces larger changes in decoded motion,
while static-region appearance changes remain smaller than whole-image
changes. Zeroing the latent actions reduces the mean per-sample motion ratio
to 0.456. Together with the qualitative examples, these measurements support
the role of latent actions in steering motion while the decoder retains much
of the scene appearance.

\noindent\begin{minipage}[t]{0.45\linewidth}
    \vspace{0pt}
    \centering
    \scriptsize
    \setlength{\tabcolsep}{2pt}
    \begin{tabular}{@{}lrrrr@{}}
        \toprule
        Conditioning & \shortstack{Motion\\ratio} & \shortstack{Flow\\deviation} & \shortstack{RGB\\L1} & \shortstack{Static\\RGB L1} \\
        \midrule
        Original & 1.000 & 0.000 & 0.0000 & 0.0000 \\
        Noise ($\alpha=0.1$) & 1.000 & 0.141 & 0.0054 & 0.0030 \\
        Noise ($\alpha=0.3$) & 0.995 & 0.211 & 0.0077 & 0.0034 \\
        Noise ($\alpha=1.0$) & 0.946 & 0.481 & 0.0215 & 0.0077 \\
        Zero queries & 0.456 & 0.866 & 0.0758 & 0.0227 \\
        \bottomrule
    \end{tabular}
    \captionof{table}{\textbf{Sensitivity to latent-action perturbations.}
    Means over eight clips and two decoder seeds. Flow deviation (pixels/frame)
    and RGB L1 are measured against unperturbed reconstructions. Motion ratio
    is averaged per clip and seed; no perturbation has ratio 1.}
    \label{tab:latent_perturbation_metrics}
\end{minipage}\hfill
\begin{minipage}[t]{0.53\linewidth}
    \vspace{0pt}
    \centering
    \includegraphics[width=\linewidth]{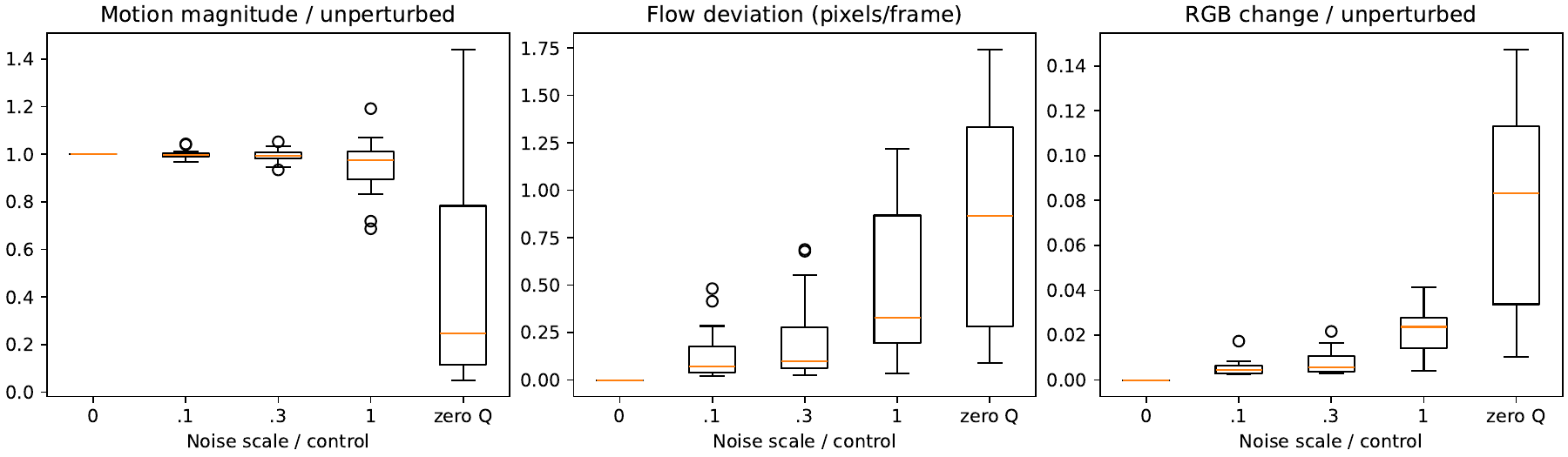}
    \captionof{figure}{\textbf{Distribution of perturbation effects.}
    Motion ratio, flow deviation, and RGB change for the same eight clips
    and two decoder seeds as Table~\ref{tab:latent_perturbation_metrics}.
    Boxes show quartiles and medians; whiskers extend to observations within
    1.5 times the interquartile range, with outliers shown individually.}
    \label{fig:latent_perturbation_boxplots}
\end{minipage}
\par\medskip

\section{Additional Qualitative Results and Analysis}
\label{app:qualitative_results}

\subsection{Any-to-Any Video Transfer}
\label{app:video_transfer_qualitative}

\subsubsection{In-domain Transfer}
\label{app:any2any}

\begin{figure}[!htbp]
    \centering
    \includegraphics[width=\textwidth]
    {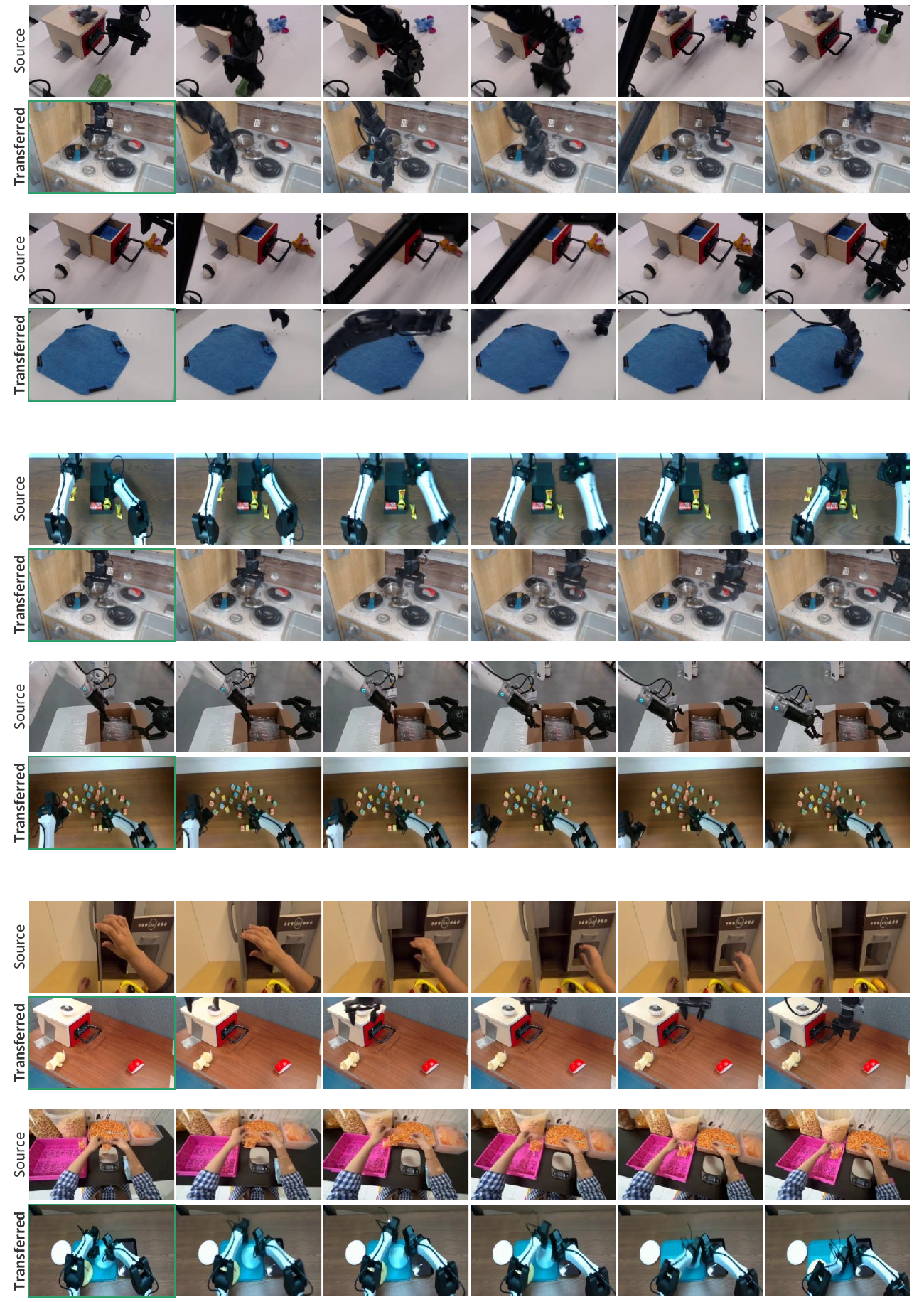}
    \caption{
    \textbf{Additional qualitative Any-to-Any video transfer results.}
    Each example shows a source motion and the corresponding transferred
    sequence generated from a target-domain initial frame.
    }
    \label{fig:any2any_additional}
\end{figure}

Figure~\ref{fig:any2any_additional} provides additional any-to-any video
transfer examples across embodiments and environments, complementing
Figure~\ref{fig:transfer_qualitative} in the main paper.

\subsubsection{Additional Zero-Shot Human-to-Robot Transfer}
\label{app:hr_transfer}

Figure~\ref{fig:hr_additional} provides additional zero-shot human-to-robot
video transfer comparisons on HumanAndRobot, complementing
Figure~\ref{fig:hr_qualitative}. All methods use the evaluation protocol in
Section~\ref{app:video_transfer_evaluation}.

\begin{figure}[!htbp]
    \centering
    \includegraphics[width=\textwidth]
    {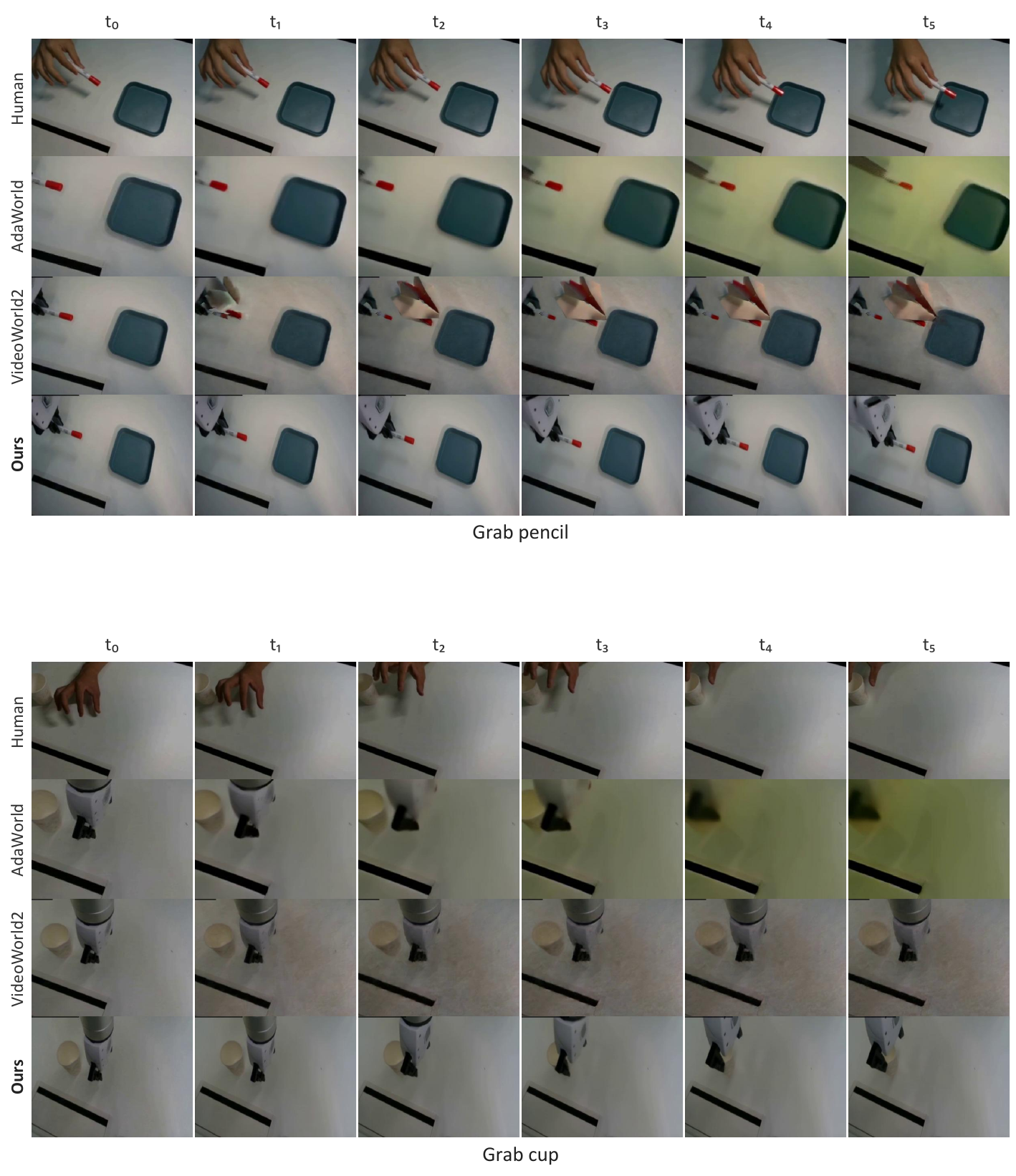}
    \caption{
    \textbf{Additional zero-shot human-to-robot video transfer results on the
    HumanAndRobot benchmark.}
    }
    \label{fig:hr_additional}
\end{figure}

\begin{figure}[!htbp]
    \ContinuedFloat
    \centering
    \includegraphics[width=\textwidth]
    {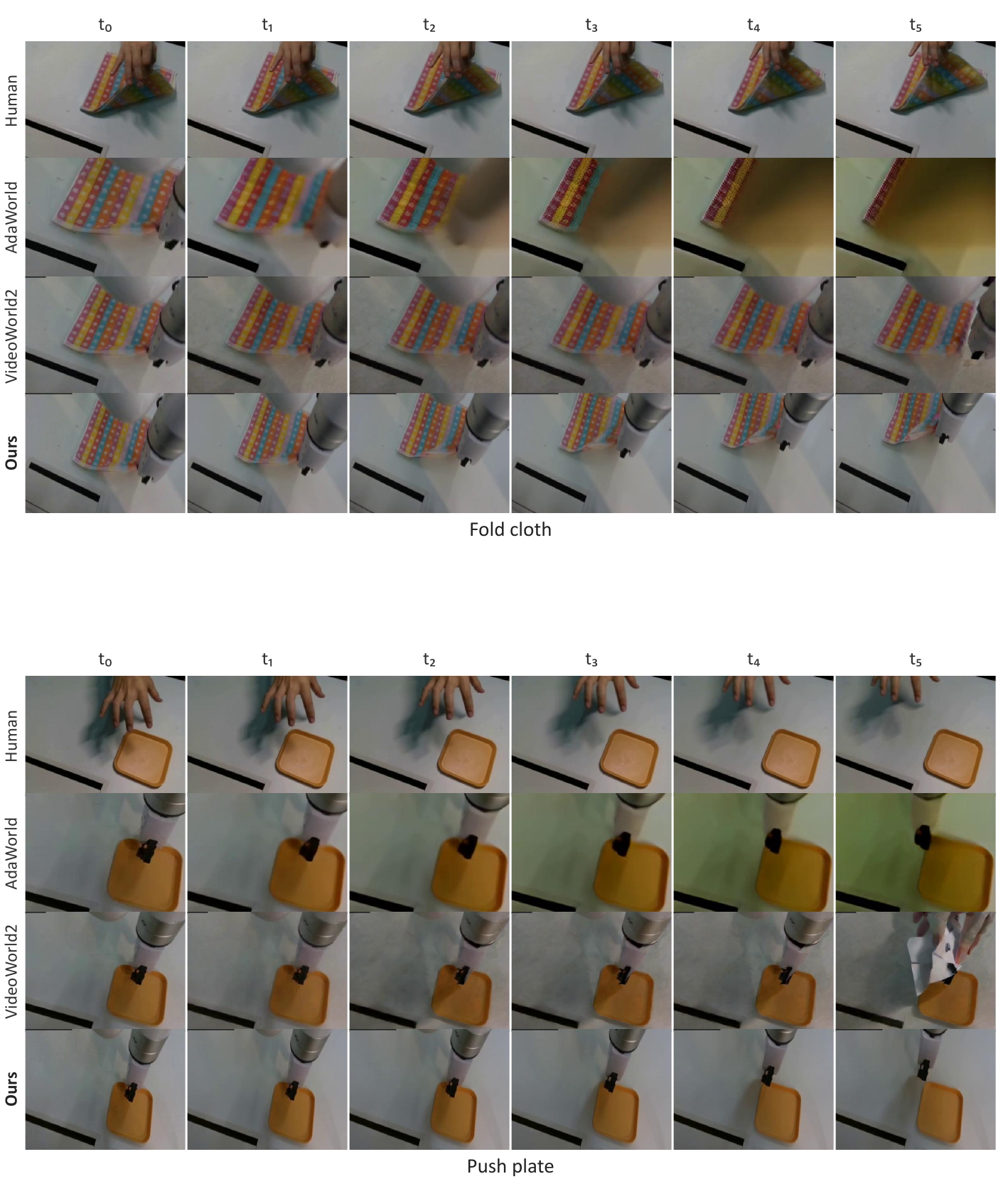}
    \caption{
    \textbf{Additional zero-shot human-to-robot video transfer results on the
    HumanAndRobot benchmark (continued).}
    }
\end{figure}

\begin{figure}[!htbp]
    \ContinuedFloat
    \centering
    \includegraphics[width=\textwidth]
    {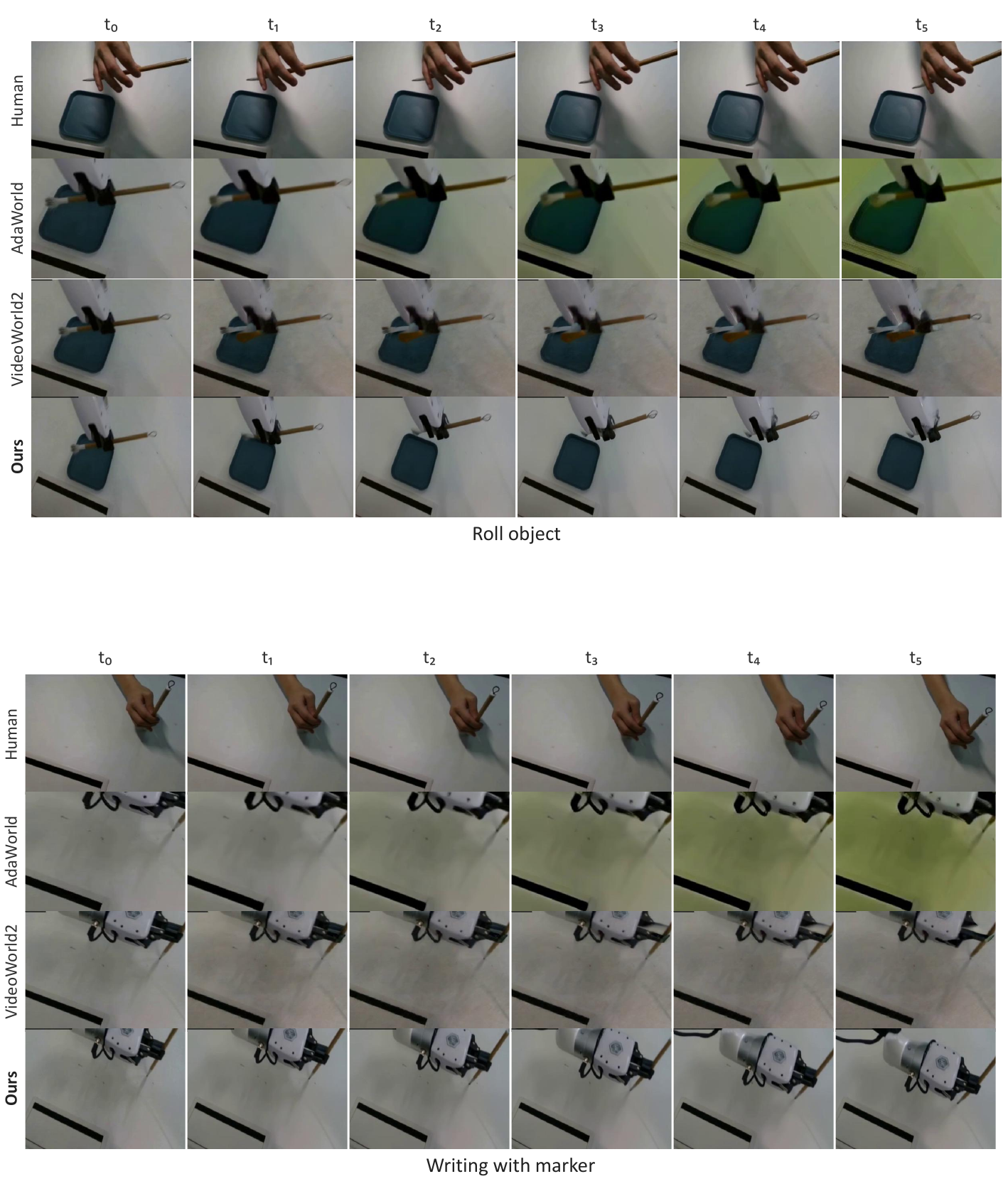}
    \caption{
    \textbf{Additional zero-shot human-to-robot video transfer results on the
    HumanAndRobot benchmark (continued).}
    }
\end{figure}

\subsection{Human-to-Robot Action Transfer}
\label{app:action_transfer_qualitative}

\begin{figure}[!t]
    \centering
    \includegraphics[width=\textwidth]
    {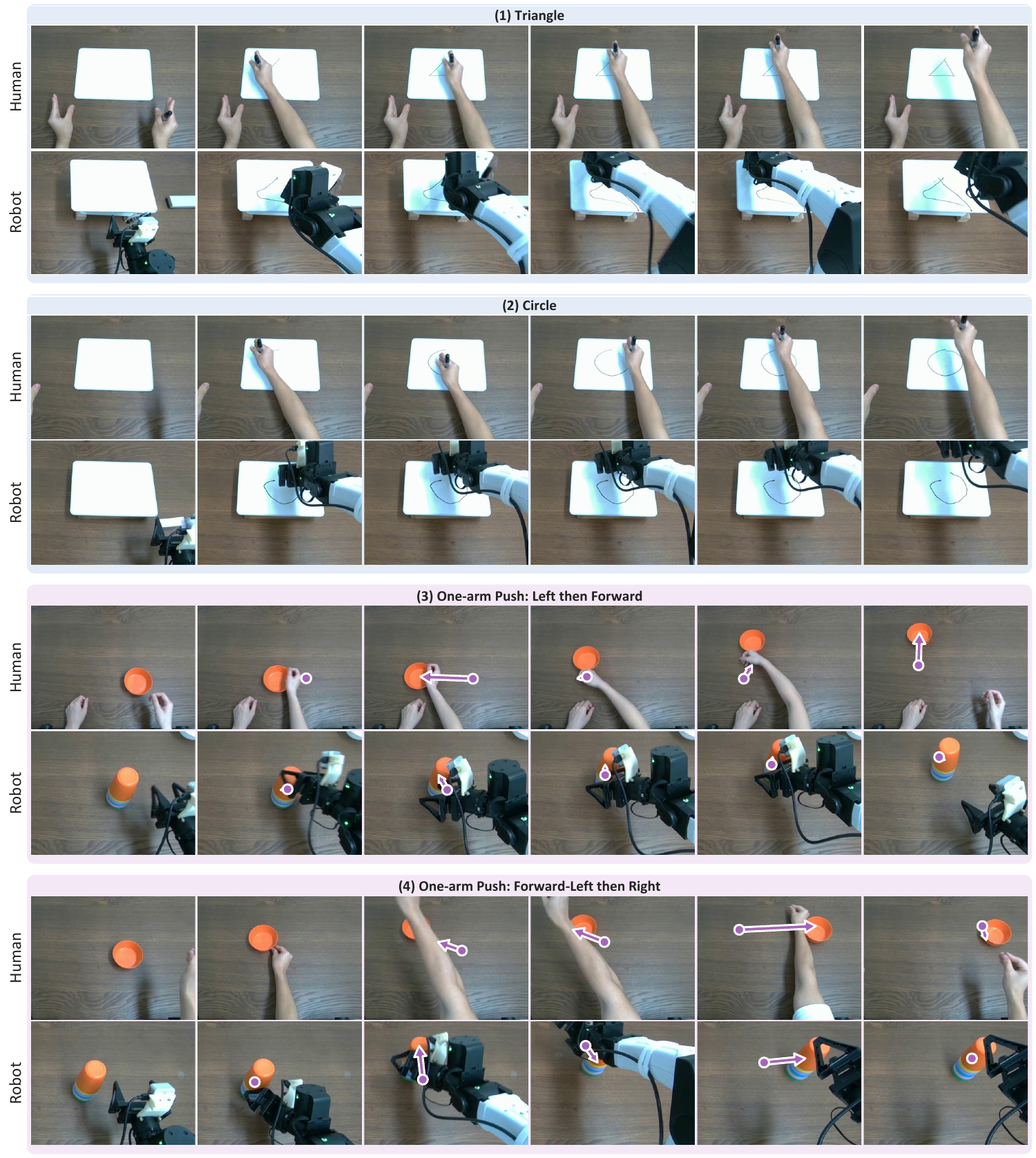}
    \caption{
    \textbf{Additional qualitative human-to-robot action transfer results.}
    DreamX transfers human demonstrations into executable robot actions while
    preserving task-specific trajectory structure and temporal ordering across
    writing and single-arm manipulation examples.
    }
    \label{fig:action_transfer_additional}
\end{figure}

Figure~\ref{fig:action_transfer_additional} provides additional human-to-robot
action transfer examples for writing and single-arm manipulation, complementing
Figure~\ref{fig:action_transfer_qualitative}. The experimental setup is described
in Section~\ref{app:action_transfer_evaluation}.

\subsection{Policy Rollouts}
\label{app:policy_qualitative}

\begin{figure}[htbp]
    \centering
    \includegraphics[width=\linewidth]{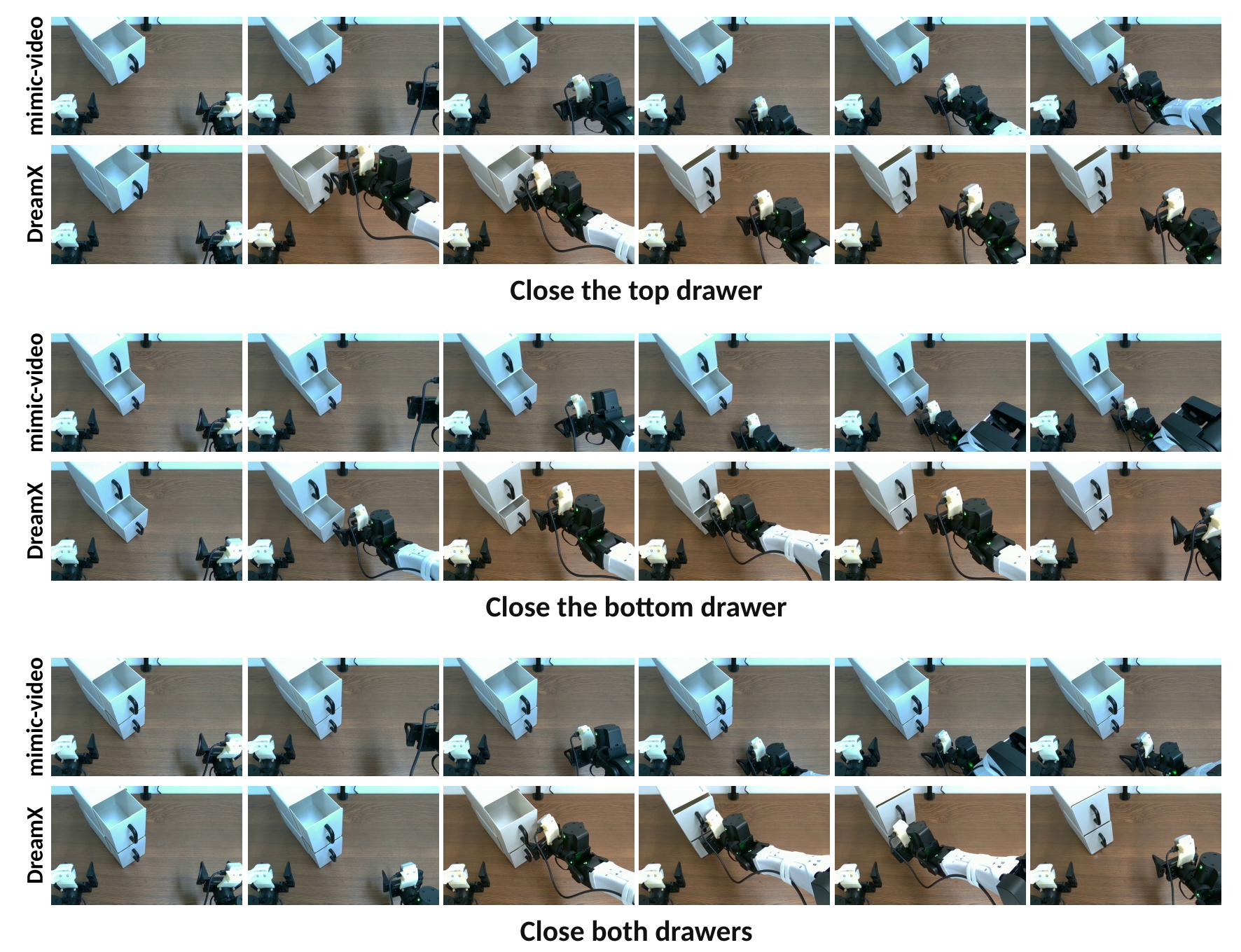}
    \caption{\textbf{Real-world drawer-closing rollouts.}
    Six frames are sampled at equal fractions of each rollout's duration.}
    \label{fig:policy_rollouts_additional}
\end{figure}

Figure~\ref{fig:policy_rollouts_additional} provides illustrative real-world
drawer-closing rollouts for the upper-only, lower-only, and both-open
configurations.

\subsection{Query-to-Video Attention}
\label{app:query_attention}

\begin{figure}[htbp]
    \centering
    \includegraphics[width=0.49\linewidth]{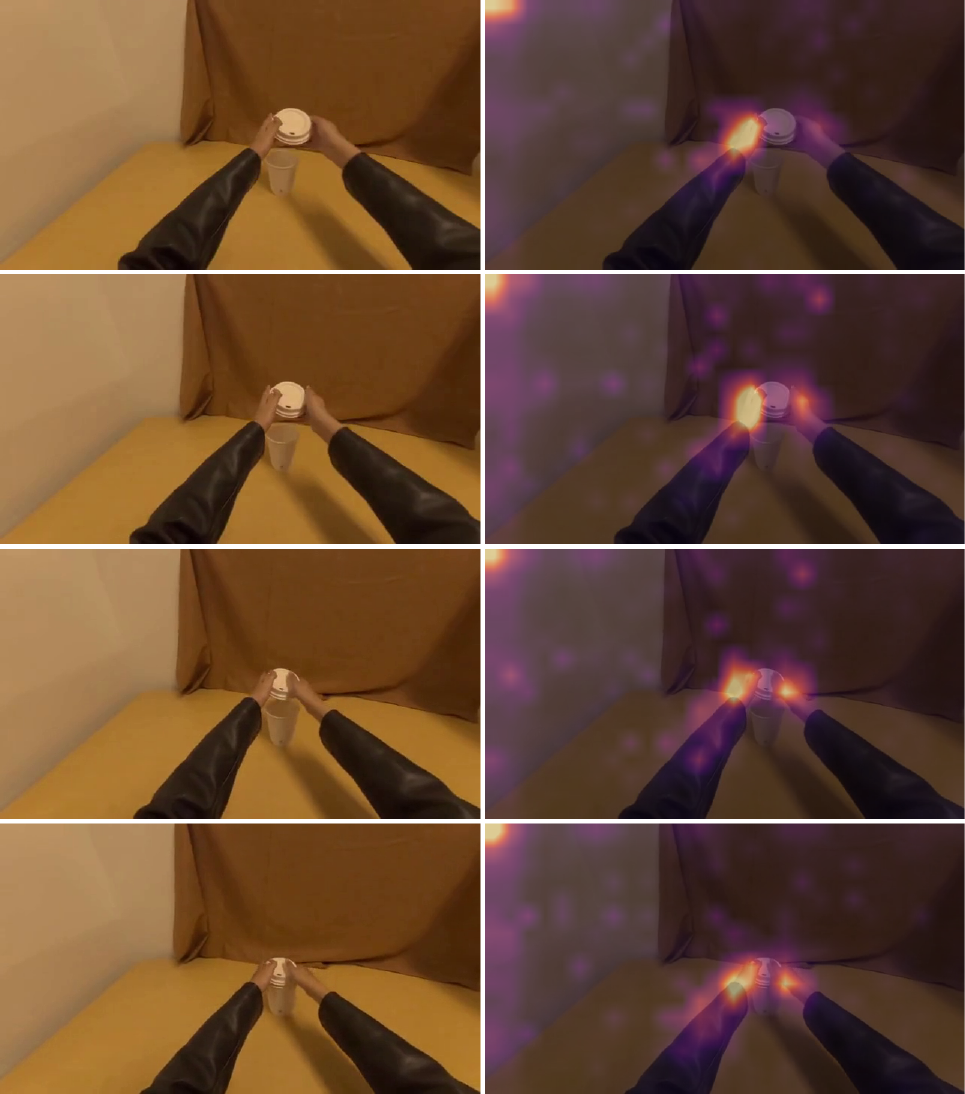}
    \hfill
    \includegraphics[width=0.49\linewidth]{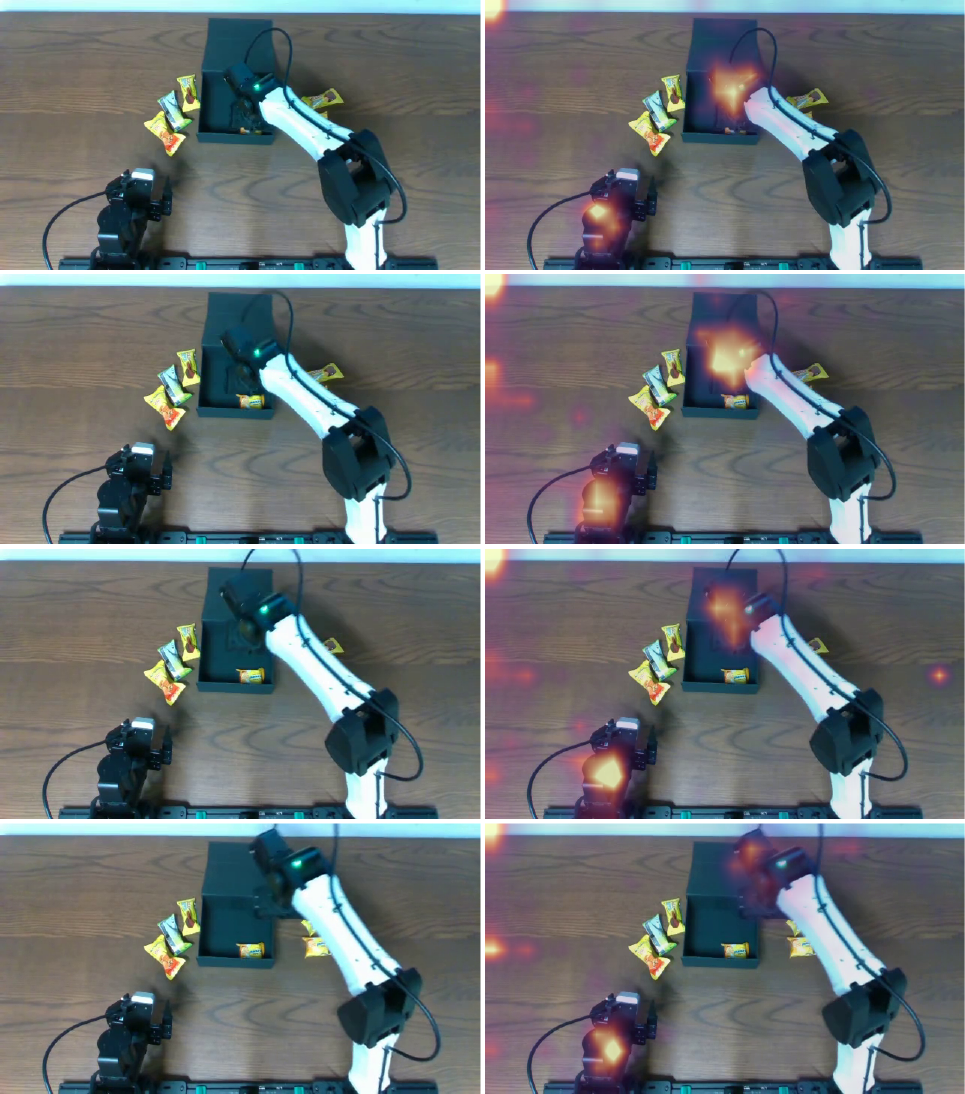}
    \caption{\textbf{Query-to-video attention across time.}
    Human demonstrations (left) and robot demonstrations (right), each shown
    with the input frame beside its attention overlay. Rows follow temporal
    order and highlight attention around the acting embodiment and objects.}
    \label{fig:attention_sequence_additional}
\end{figure}

Figure~\ref{fig:attention_sequence_additional} provides additional
query-to-video attention visualizations across human and robot demonstration
sequences, complementing Figure~\ref{fig:query_attention}. We average query-to-video
attention over all heads, the four query slots within each temporal group,
and the final seven transformer blocks.

% \section{Limitations and Future Work}
% \label{app:limitations}

% DreamX provides a common latent-action interface across embodiments, but
% transferring motion does not by itself guarantee successful physical
% interaction. Our action-transfer experiments execute predicted commands
% open-loop, so they cannot correct errors caused by object displacement or
% imperfect contact. Incorporating visual and tactile feedback is a promising
% direction for contact-sensitive and long-horizon transfer. Our real-world
% policy evaluation focuses on drawer closing with one bimanual robot;
% evaluating broader tasks, robot morphologies, and environments would further
% establish the scope of the approach. Finally, although compact latent actions
% avoid future-video generation during policy rollout, inference still relies
% on iterative denoising. Distillation and more efficient sampling could further
% reduce latency for closed-loop control.

\end{document}